\documentclass{article}

\PassOptionsToPackage{numbers, compress}{natbib}
 \usepackage[preprint]{neurips_2026}

\usepackage[utf8]{inputenc} 
\usepackage[T1]{fontenc}    
\usepackage{hyperref}       
\usepackage{url}            
\usepackage{booktabs}       
\usepackage{amsfonts}       
\usepackage{nicefrac}       
\usepackage{microtype}      
\usepackage{xcolor}         

\usepackage{graphicx}
\usepackage{kotex}
\usepackage{amsmath}
\usepackage{multirow}
\usepackage{float} 
\usepackage{subcaption}

\newcommand{\graycell}[1]{\textcolor{gray}{#1}}
\definecolor{DeltaRed}{RGB}{200,0,0}
\definecolor{DeltaBlue}{RGB}{0,70,200}
\newcommand{\red}[1]{\textcolor{DeltaRed}{#1}}
\newcommand{\blue}[1]{\textcolor{DeltaBlue}{#1}}

\title{Rank-factorized Implicit Neural Bias: Scaling Super-Resolution Transformer with FlashAttention}

\author{%
  Dongheon Lee \\
  University of Seoul\\
  \texttt{dslisleedh@uos.ac.kr} \\
  \And
  Seokju Yun \\
  KAIST AI\\
  \texttt{sj\_yun@kaist.ac.kr} \\
  \And
  Jaegyun Im \\
  University of Seoul\\
  \texttt{imij0522@uos.ac.kr} \\
  \And
  Youngmin Ro\thanks{Corresponding author} \\
  University of Seoul\\
  \texttt{youngmin.ro@uos.ac.kr} \\
  \small{Code: \url{https://github.com/dslisleedh/SST}}
}

\begin{document}

\maketitle

\begin{figure}[H]
  \centering
  \includegraphics[width=0.9\textwidth]{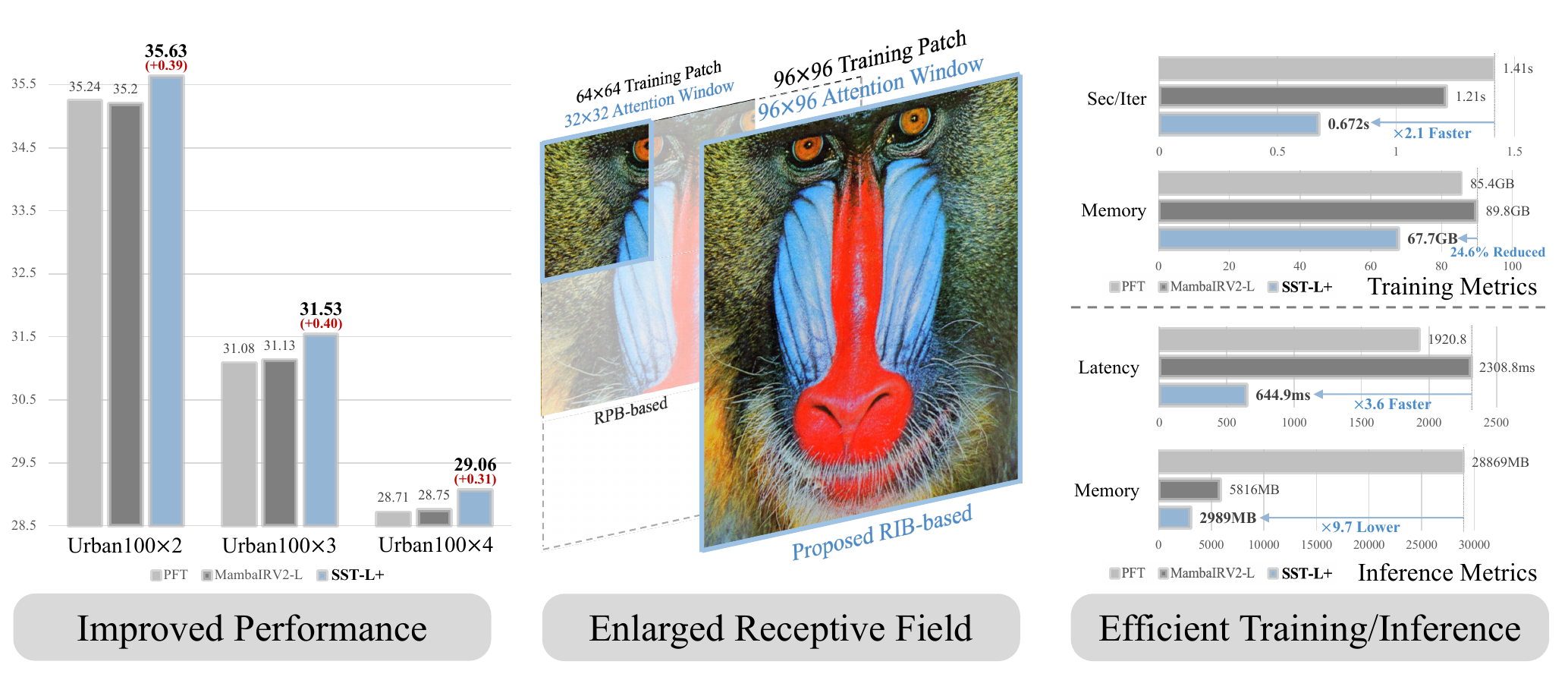}
  \caption{
    Overview of our contributions. 
    By replacing relative positional bias (RPB) with the proposed Rank-factorized Implicit Neural Bias (RIB), we enable FlashAttention in SR Transformers. 
    The resulting efficiency allows us to scale the attention window, training patch size, and training data, leading to improved performance with significantly lower training and inference costs.
  }
  \label{fig:teaser}
\end{figure}

\begin{abstract}
Recent Super-Resolution~(SR) methods have increasingly adopted Transformers for their long-range modeling capability and high representational capacity.
However, their reliance on relative positional bias~(RPB), typically injected as an additive term in attention logits, limits the direct use of hardware-efficient attention kernels such as FlashAttention.
This limitation imposes a prohibitive computational burden during both training and inference, preventing SR Transformers from scaling to larger attention windows or training patches.
In this paper, we propose Rank-factorized Implicit Neural Bias~(RIB), an alternative to RPB that enables FlashAttention in SR Transformers.
Specifically, RIB encodes positional bias as rank-factorized implicit features and concatenates them with input tokens, reformulating biased attention into a standard dot-product form compatible with FlashAttention.
Further, we introduce convolutional local attention and a cyclic window strategy to better exploit the larger attention windows enabled by RIB and FlashAttention.
These designs are integrated into Scalable SR Transformer~(SST), a new SR Transformer that preserves the modeling power of biased attention while making it compatible with FlashAttention.
By substantially reducing the training and inference costs compared to the RPB-based SR Transformers, SST enables much larger training patches, attention windows, and datasets, achieving substantial performance gains.
\end{abstract}

\section{Introduction}
\vspace{-0.1cm}
Super-Resolution~(SR) aims to reconstruct a high-resolution~(HR) image from a low-resolution~(LR) input and has long been a fundamental problem in computer vision. 
Modeling long-range dependencies is crucial for SR, as distant but correlated patterns, such as repeated textures and edges, can help disambiguate the inherently ill-posed LR inputs.
Accordingly, Transformers~\cite{Transformers} have recently emerged as a promising architecture because their core operator, self-attention, enables long-range interactions and input-adaptive feature aggregation, offering strong representational capacity.

However, adopting Transformers for SR in practice still faces several limitations that prevent fully exploiting their representational capacity:
(i) \emph{The quadratic cost of self-attention} in the number of tokens makes global attention prohibitively expensive for high-resolution feature maps that consist of pixel-level tokens without a patchify stem. 
As a result, many SR Transformers resort to window-based attention with relatively small windows~\cite{SwinTransformer, SwinIR}, which restricts the capturing of long-range dependencies across distant but correlated regions.
(ii) \emph{Most methods are limited to training on small cropped patches}~(e.g., $64\times64$), since leveraging larger patches~(e.g., $96\times96$) substantially increases the training budget, although the richer global context can improve performance~\cite{EffectiveLongContextScaling, LSDIR, ESC}.
(iii) \emph{Many SR Transformers are still trained on relatively small datasets} such as DF2K~\cite{DIV2K, DF2KDataset} (3,450 images), despite the availability of much larger datasets like LSDIR~\cite{LSDIR} (84,991 images) and DiverSeg-IP~\cite{DiverSeg} (526,503 images), unlike other vision domains~\cite{ViT, DINOV1, DINOV2, SAM1}.

Currently, many SR methods primarily mitigate the limitation (i), while the orthogonal gains from limitations (ii) and (iii) remain underexplored.
We identify SR Transformers’ heavy reliance on Relative Positional Bias~(RPB) as a key, yet often overlooked, bottleneck that limits the use of hardware-efficient attention kernels in SR Transformers.
RPB injects a spatial prior into self-attention by adding a learnable, distance-dependent bias to the score matrix, and is crucial for SR performance.
In practice, incorporating RPB typically requires either materializing the score matrix or additional indexing and memory reads, making it difficult to use with hardware-efficient self-attention kernels~(e.g., FlashAttention~\cite{FlashAttn, FlashAttn2, FlashAttn3}), which aim to avoid materializing the score matrix and to reduce memory I/O.
This incompatibility forces existing models to rely on slow, memory-intensive implementations, which in turn hinders the effective scaling of training patch size and data. 
Consequently, instead of pursuing scalability, previous works have been confined to designing complex windowing strategies~\cite{CAT, ART, DAT, HiTSR} or sub-quadratic alternatives, such as linear-complexity attention~\cite{Restormer, RGT, DAT, HiTSR}, Mamba~\cite{MambaIR, MambaIRV2, TSPMamba}, and sparse operators~\cite{PFT}, to bypass this efficiency wall.

In this paper, we propose \textbf{R}ank-factorized \textbf{I}mplicit Neural \textbf{B}ias~(RIB), a FlashAttention-compatible alternative to conventional RPB that enables SR Transformers to scale efficiently.
Specifically, RIB parameterizes positional bias into two low-rank implicit neural representations generated by a coordinate-based multi-layer perceptron~(MLP).
Then, the resulting spatial representations are injected into self-attention by concatenating them to pixel tokens, thereby memory-efficiently emulating the element-wise bias addition through dot products.
This novel design separates the number of bias parameters from the window size and remains compatible with any FlashAttention implementations, unlike traditional RPB.
Moreover, our RIB preserves the integrity of pixel representations compared to Rotary Positional Embedding~(RoPE)~\cite{RoPE, RoPEViT}, which injects spatial prior into self-attention via rotating pixel tokens.
Finally, to better exploit the larger window self-attention enabled by our RIB and FlashAttention, we propose a Convolutional Local Attention~(CLA) to focus self-attention on a larger context and a cyclic window strategy that periodically expands the attention window to extract higher-order multi-scale features in a scan-and-focus manner.

Building on these components, we introduce a FlashAttention-accelerated SR Transformer, dubbed \textbf{S}calable \textbf{S}R \textbf{T}ransformer~(SST).
By replacing conventional RPB with RIB, SST scales attention windows and training patches up to \textbf{96$\times$96}, significantly improving the performance--efficiency trade-off.
As shown in Figure~\ref{fig:teaser}, a 20M-parameter SST trained on DFLIP achieves a substantial $+$\textbf{0.4\,dB PSNR} gain on Urban100$\times3$ over previous SOTA methods~\cite{PFT} trained on the same data.
Despite its larger $96\times96$ windows and patches, SST trains $\mathbf{2.1}\times$ \textbf{faster} with \textbf{24.6\% lower memory} than prior methods~\cite{MambaIRV2} using $64\times64$ patches, while also greatly reducing inference latency and memory usage.
Comprehensive ablations and visual analyses verify that RIB, CLA, and the cyclic window strategy each contribute to SST's scalability, efficiency, and reconstruction quality.

In summary, our contributions are threefold:
(1) We identify conventional RPB as an overlooked efficiency bottleneck in SR Transformers and propose RIB, a FlashAttention-compatible positional bias that efficiently emulates element-wise relative bias via factorized implicit neural representations.\\
(2) We introduce SST, equipped with CLA and a cyclic window strategy, which scales attention windows and training patches to 96$\times$96 and training data to 614,944 images while substantially reducing training and inference costs.
(3) With joint scaling across window size, patch size, and training data, SST-L$+$ achieves state-of-the-art performance, reaching 35.63\,dB PSNR on Urban100$\times$2.
\begin{figure}[t]
  \centering
  \includegraphics[width=\textwidth]{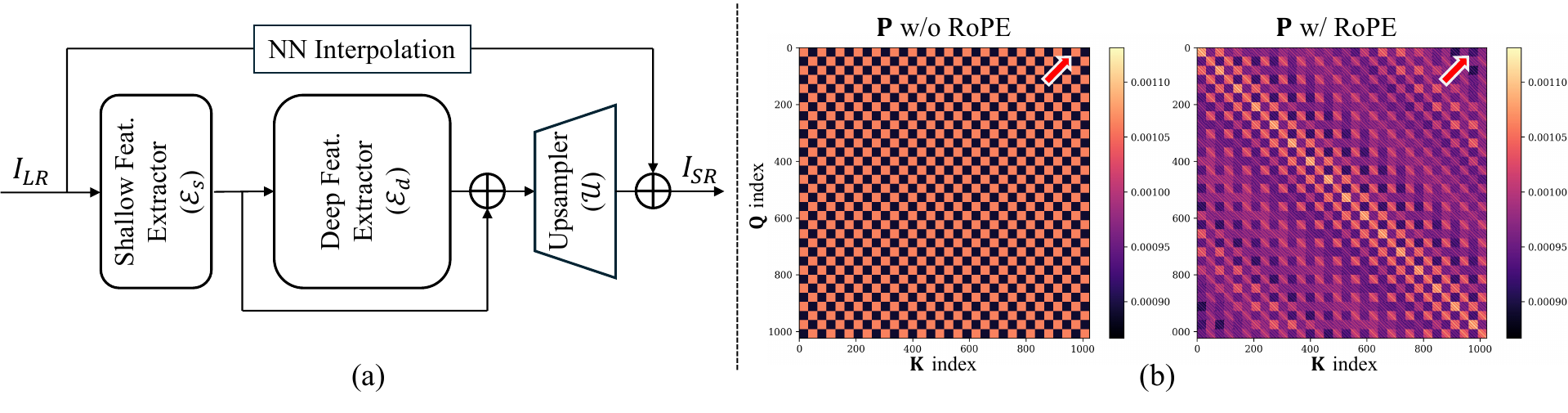}
  \vspace{-0.5cm}
  \caption{
    Preliminaries and motivation. (a) Overview of the general SR architecture. (b) Toy illustration of how RoPE affects repeated-pattern attention. We construct a 32$\times$32 feature map by repeatedly tiling two orthogonal 32-dimensional content vectors and set $\mathbf{Q}=\mathbf{K}$. 
    Without RoPE, tokens with the same repeated content maintain high mutual attention regardless of their spatial offset, yielding a checkerboard pattern. 
    With RoPE, however, the dot product between repeated content vectors is modulated by relative-position-dependent phases. 
    Therefore, the same repeated patterns can receive different attention probabilities depending only on their offset, as highlighted by the red arrow.
  }
  \vspace{-0.5cm}
  \label{fig:preliminaries}
\end{figure}

\vspace{-0.1cm}
\section{Preliminaries}
\vspace{-0.1cm}
In this section, we briefly review the key components relevant to our method,
including general SR architectures, SR Transformers, window-based self-attention, positional bias, and FlashAttention. 

\vspace{-0.1cm}
\subsection{SR Architecture}
\vspace{-0.1cm}
As shown in Figure~\ref{fig:preliminaries}~(a), we follow the standard residual-in-residual SR pipeline~\cite{RCAN}, which consists of a shallow feature extractor $\mathcal{E}_s$, a deep feature extractor $\mathcal{E}_d$, and an upsampler $\mathcal{U}$.
Given an LR image $I_{\mathrm{LR}}\in\mathbb{R}^{H\times W\times 3}$, where $H$ and $W$ denote its height and width, $\mathcal{E}_s$ first extracts shallow features $F_s\in\mathbb{R}^{H\times W\times D}$, where $D$ is the channel dimension.
The deep feature extractor $\mathcal{E}_d$ then refines $F_s$ into $F_d\in\mathbb{R}^{H\times W\times D}$, and $\mathcal{U}$ reconstructs an SR image from $F_s+F_d$.
We additionally adopt a nearest-neighbor image skip connection~\cite{ABPN}, treating $\mathcal{U}$ as a residual image predictor~\cite{VDSR}, so the final output is $I_{\mathrm{SR}}\in\mathbb{R}^{rH\times rW\times 3}$ for an upscaling factor $r$.
Since this backbone is standard, the rest of this section focuses on the self-attention operator used inside $\mathcal{E}_d$.

\vspace{-0.1cm}
\subsection{SR Transformers and Window-based Self-attention} 
\vspace{-0.1cm}
Given token features $X\in\mathbb{R}^{N\times D}$, where $N$ denotes the number of tokens, self-attention provides input-adaptive global feature aggregation, $\mathbf{O}=\mathbf{PV}$, with $\mathbf{S}=\mathbf{QK}^\top/\sqrt{D}$ and $\mathbf{P}=\mathrm{SoftMax}(\mathbf{S})$ as the $N\times N$ attention score and probability matrices.
While this input-adaptive global aggregation provides strong representational capability, it requires quadratic complexity $\mathcal{O}(N^2D)$ with respect to the token numbers, which becomes prohibitive in SR. 
Unlike high-level vision tasks that tokenize images into coarse patches, SR commonly treats each pixel as a token to preserve fine details, yielding significantly higher $N=HW$.
For example, a Vision Transformer~(ViT)~\cite{ViT} with a $224\times224$ input and a $16\times16$ patchify stem uses $14^2=196$ patch tokens, or $197$ tokens including a \texttt{[CLS]} token.
Whereas a $\times2$ SR Transformer for a $1280\times720$ output processes $640\times360=230{,}400$ LR pixel tokens.
To make self-attention practical on such dense feature maps, most SR Transformers compute attention within local $M\times M$ windows~(e.g., $\{8, 16, 32\}$), reducing the complexity to $\mathcal{O}(NM^2D)$.
Although efficient, this local restriction weakens direct long-range interactions between distant but correlated patterns, making the attention window size a key bottleneck for scaling SR Transformers.

\vspace{-0.1cm}
\subsection{FlashAttention and Positional Bias}
\vspace{-0.1cm}
In large language models~(LLMs)~\cite{LLaMa3, Qwen3}, hardware-efficient attention kernels such as FlashAttention~\cite{FlashAttn, FlashAttn2, FlashAttn3, FlashAttn4} are widely adopted to mitigate the memory bottlenecks of long-context self-attention, as these models often process massive numbers of tokens.
Specifically, FlashAttention leverages IO-aware tiling and kernel fusion to compute exact self-attention without materializing the full $N\times N$ matrices~($\mathbf{S}$ and $\mathbf{P}$), thereby reducing memory traffic.
These optimizations significantly accelerate both training and inference speed compared to the standard implementations, and have been instrumental in scaling Transformers.

However, FlashAttention is not directly applicable to many SR Transformers, as they incorporate relative position bias (RPB)~\cite{T5, SwinTransformer} to inject positional priors into self-attention, often achieving notable performance gains.
Technically, RPB adds a distance-dependent learnable bias matrix $\mathbf{B}\in\mathbb{R}^{1\times N\times N}$ to the score matrix, i.e., $\tilde{\mathbf{S}} = \mathbf{S} + \mathbf{B}$.
Therefore, incorporating RPB either requires explicitly materializing an additional $N\times N$ matrix or performing extra memory accesses and indexing into a bias table to add the corresponding biases~\cite{FlexAttn, ESC}.
Such operations break the assumptions behind many fused FlashAttention kernels, and therefore are not supported in a large portion of existing FlashAttention implementations.

In contrast, LLMs often adopt a RoPE~\cite{RoPE} for the positional prior, which is fully compatible with FlashAttention.
RoPE injects positional prior by applying distance-dependent rotations to $\mathbf{Q}/\mathbf{K}$, so that relative phase differences are reflected during the dot-product computation of $\mathbf{S}$.
Since RoPE does not require materializing a $N\times N$ matrix or additional memory movement, it can be seamlessly used with FlashAttentions, and recent works have also explored its adaptation to ViT~\cite{RoPEViT}.
Nevertheless, as shown in Figure~\ref{fig:preliminaries}~(b), RoPE can substantially weaken similarities between repeated patterns at large spatial offsets, which is crucial for the SR task, due to phase wrap/aliasing, especially under small head dimensions~(e.g., $\approx\!32$).
Moreover, since RoPE injects positional prior by directly modulating the dot product with rotated $\mathbf{Q}/\mathbf{K}$, the positional effect is entangled with the pixel representations.
Consequently, compensating for undesired distance-dependent suppression may require altering the $\mathbf{Q}/\mathbf{K}$ content representations themselves, consuming fitting capacity and potentially degrading performance.
This motivates us to introduce a FlashAttention-compatible attention bias that explicitly decouples the positional prior from the pixel representation.

\vspace{-0.1cm}
\section{Proposed Methods}
\vspace{-0.1cm}
In this section, we describe our proposed methods, including a Rank-factorized Implicit Neural Bias~(RIB), a Convolutional Local Attention~(CLA), and a cyclic window strategy.

\begin{figure}[t]
  \centering
  \includegraphics[width=\textwidth]{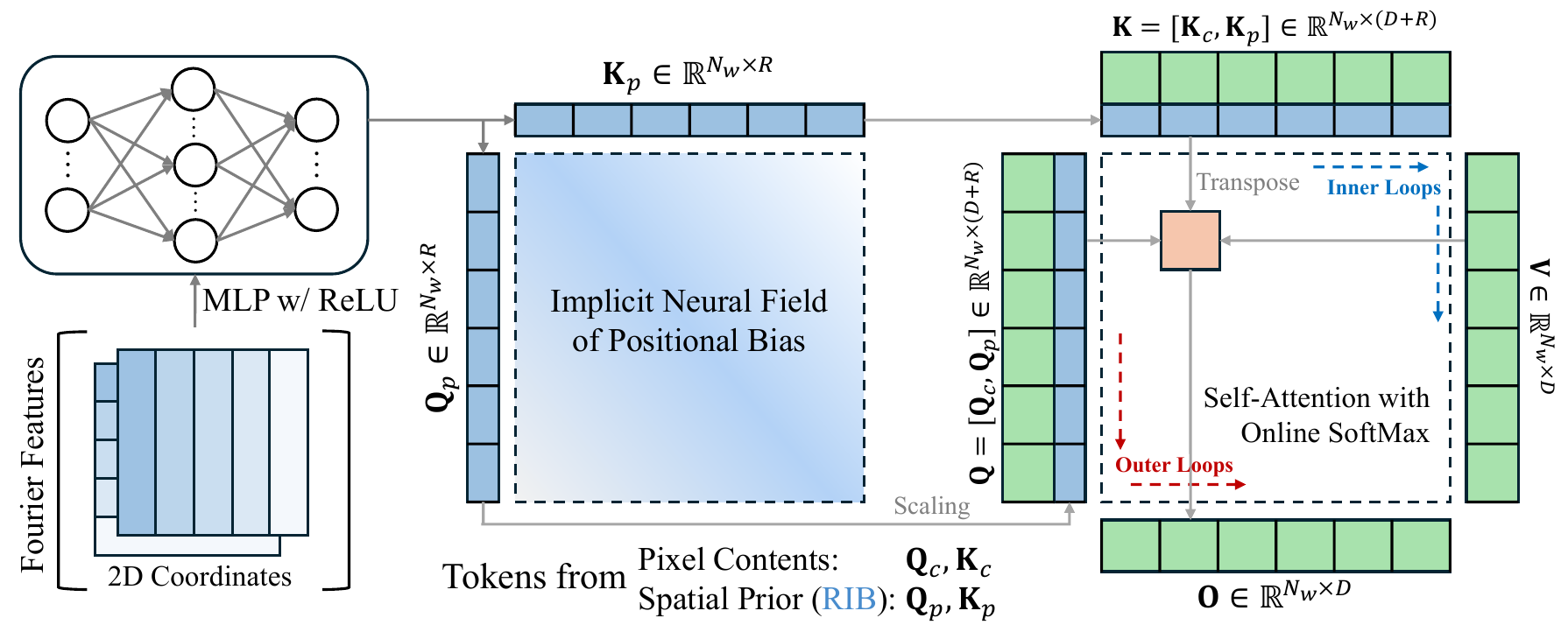}
  \vspace{-0.5cm}
  \caption{
    Overall illustration for proposed Rank-factorized Implicit Neural Bias~(RIB).
  }
  \label{fig:RIN}
  \vspace{-0.5cm}
\end{figure}

\subsection{Rank-factorized Implicit Neural Bias}
The main goal of the proposed RIB is to provide a positional prior for the self-attention's $\mathbf{S}$, while explicitly disentangling pixel content from positional priors and maintaining full compatibility with FlashAttention.
To this end, we approximate positional bias using per-token 2D coordinates and their Fourier features.
For an $M \times M$ attention window, let $N_w=M^2$ denote the number of tokens within the window.
We assign each token a normalized 2D coordinate $\mathbf{x}\in[-1,1]^{N_w\times 2}$.
Then, we augment these coordinates with a Fourier feature mapping~\cite{NERF} to obtain the coordinate embedding $\mathbf{r}_{\mathrm{in}}\in\mathbb{R}^{N_w\times (2+4L)}$, which is defined as:
\begin{equation}
\mathbf{r}_{\mathrm{in}}=\gamma(\mathbf{x})
=\Big[\,\mathbf{x},\ \sin\!\big(2^{0}\mathbf{x}\big),\ \cos\!\big(2^{0}\mathbf{x}\big),\ \ldots,\ 
\sin\!\big(2^{L-1}\mathbf{x}\big),\ \cos\!\big(2^{L-1}\mathbf{x}\big)\Big],
\end{equation}
where $[\cdot]$ denotes concatenation along the feature dimension, and $\sin(\cdot)$ and $\cos(\cdot)$ are applied element-wise.
The $L$ is the number of frequency bands in the positional encoding.
Next, we feed $\mathbf{r}_{\mathrm{in}}$ into a lightweight MLP with ReLU activation to produce factorized positional representations that parameterize an implicit neural field for positional bias.
Specifically, we first compute a shared hidden representation
$\mathbf{h}\in\mathbb{R}^{N_w\times d_h}$ as:
\begin{equation}
\mathbf{h}=\mathrm{ReLU}\!\left(\mathbf{r}_{\mathrm{in}}\mathbf{W}_{h}+\mathbf{b}_{h}\right),
\end{equation}
where $\mathbf{W}_{h}\in\mathbb{R}^{(2+4L)\times d_h}$ and $\mathbf{b}_{h}\in\mathbb{R}^{d_h}$.
We then linearly project $\mathbf{h}$ into a low-rank space of dimension $R$, as follows:
\begin{equation}
\mathbf{Q}_{\mathrm{p}}=\mathbf{h}\mathbf{W}_{p,q},\qquad
\mathbf{K}_{\mathrm{p}}=\mathbf{h}\mathbf{W}_{p,k},
\end{equation}
where $\mathbf{W}_{p,q},\mathbf{W}_{p,k}\in\mathbb{R}^{d_h\times R}$. 
In practice, we share the hidden representation \(\mathbf{h}\) across all attention heads, and obtain head-wise \(\mathbf{Q}_{\mathrm{p}}\) and \(\mathbf{K}_{\mathrm{p}}\) via separate linear projections from the shared \(\mathbf{h}\).
Next, unlike vanilla self-attention that directly forms $\mathbf{Q}=X\mathbf{W}_q$ and $\mathbf{K}=X\mathbf{W}_k$, we explicitly separate content and positional components.
Given token features $X$, we define the content projections
\begin{equation}
\mathbf{Q}_{\mathrm{c}}=X\mathbf{W}_{q},\qquad
\mathbf{K}_{\mathrm{c}}=X\mathbf{W}_{k},
\end{equation}
while keeping the value projection identical to standard attention~($\mathbf{V}=X\mathbf{W}_{v}$).
Then, we merge content and positional components by channel-wise concatenation.
Specifically, we scale $\mathbf{Q}_{\mathrm{c}}$ and $\mathbf{Q}_{\mathrm{p}}$ by their respective dimensions and construct
\begin{equation}
\mathbf{Q}=\Big[\ \mathbf{Q}_{\mathrm{c}}/\sqrt{D}\ ,\ \mathbf{Q}_{\mathrm{p}}/\sqrt{R}\ \Big],\qquad
\mathbf{K}=\Big[\ \mathbf{K}_{\mathrm{c}}\ ,\ \mathbf{K}_{\mathrm{p}}\ \Big].
\label{eq:rin_qk_concat}
\end{equation}
The resulting scaled similarity matrix is obtained as a single dot-product in the augmented channel space:
\begin{equation}
\mathbf{S}=\mathbf{Q}\mathbf{K}^{\top}
=
\underbrace{\bigl(\mathbf{Q}_{\mathrm{c}}\mathbf{K}_{\mathrm{c}}^{\top}\bigr)/\sqrt{D}}_{\text{content term}}
+
\underbrace{\bigl(\mathbf{Q}_{\mathrm{p}}\mathbf{K}_{\mathrm{p}}^{\top}\bigr)/\sqrt{R}}_{\text{bias term}},
\label{eq:rin_score}
\end{equation}
which shows that the spatial prior is injected as an additive bias term in the logits while remaining decoupled from the pixel content term in logit space. 
The subsequent steps operate identically to standard self-attention~(e.g., $\mathbf{O}=\mathrm{SoftMax}(\mathbf{S})\mathbf{V}$).

In summary, RIB is fully compatible with existing FlashAttention implementations, similar to RoPE, because it injects positional priors by augmenting \(\mathbf{Q}\) and \(\mathbf{K}\) and computing the logits via a single dot-product, without requiring any extra \(M\times M\) bias matrix materialization or table indexing.
Moreover, unlike RPB, whose number of bias parameters grows with the window size~(e.g., \((2M-1)^2 \approx O(M^2)\)), RIB parameterizes the positional bias using a lightweight MLP, making the number of bias parameters independent of the window size (e.g., \(O(d_h(L+R))\)).
Next, unlike FlashBias~\cite{FlashBias}, which post-hoc factorizes learned biases for acceleration, RIB natively parameterizes attention bias during training via coordinate-conditioned low-rank features, providing an explicit spatial prior without relying on a pre-existing RPB table or learned representations.
Finally, since \(\mathbf{Q}_{\mathrm{p}}\) and \(\mathbf{K}_{\mathrm{p}}\) depend only on the window geometry and not on the input pixel tokens, they can be precomputed and cached, further reducing inference-time overhead.

\subsection{Convolutional Local Attention}
Due to the low-rank nature of the RIB, it may be less effective for certain highly localized, rapidly varying positional patterns.
We therefore further enhance RIB with CLA, which strengthens short-range interactions and helps capture fine details.
Given token features $X$, we define a pair of reshaping operators $\mathcal{F}$ and $\mathcal{F}^{-1}$ that convert between token features and 2D feature maps:
\begin{equation}
\mathcal{F}:\mathbb{R}^{H\times W\times D}\rightarrow\mathbb{R}^{N\times D},
\qquad
\mathcal{F}^{-1}:\mathbb{R}^{N\times D}\rightarrow\mathbb{R}^{H\times W\times D}.
\end{equation}
CLA computes a local gate by applying convolutions to the restored 2D feature map:
\begin{equation}
\tilde{\mathbf{G}} =
\sigma\!\left(
\mathrm{PWConv}\!\left(
\mathrm{DWConv}_{3\times3}\!\left(\mathcal{F}^{-1}(X)\right)
\right)
\right)
\in\mathbb{R}^{H\times W\times D}.
\end{equation}
where $\sigma(\cdot)$ denotes the sigmoid activation, and $\odot$ denotes element-wise multiplication.
After being converted back to token form, this gate modulates the self-attention output:
\begin{equation}
\mathbf{O}'=\mathbf{O}\odot\mathbf{G},
\quad
\mathbf{G}=\mathcal{F}(\tilde{\mathbf{G}})\in\mathbb{R}^{N\times D}.
\end{equation}

In summary, CLA selectively emphasizes spatially consistent high-frequency responses, aiding self-attention with RIB to focus on broader and structural patterns.
Moreover, compared to the output projection of standard self-attention~($Y=\mathbf{O}\mathbf{W}_{o}$), where $\mathbf{W}_{o}$ denotes output projection weight matrix, CLA provides additional non-linearity before output projection~($Y=(\mathbf{O}\odot\mathbf{G})\mathbf{W}_{o}$), further enhancing the representational capability of Transformers~\cite{GatedAttn}.

\subsection{Cyclic Window Strategy}
Thanks to the hardware-efficiency enabled by RIB and FlashAttention, we can employ larger attention windows without prohibitive overhead. 
However, using only large windows is not always optimal, as mixing window sizes benefits multi-scale feature extraction in SR~\cite{ELAN, HiTSR}.
We therefore adopt a cyclic window strategy that recycles local-to-broad aggregation~(e.g., $\{16, 32, 64, 16, 32, 64\}$), rather than expanding the window size only once. 
Intuitively, the early cycle scans contexts to enrich features with long-range structures, while the later cycle performs more focused aggregation over the enriched features. 
This scan-and-focus behavior distinguishes our strategy from a single
monotonic increase~\cite{HiTSR}, and is further supported by the positional bias visualization in Figure~\ref{fig:PB}.

\section{Experiments}
We evaluate our method in terms of (i) training and inference efficiency, (ii) quantitative SR performance, (iii) ablations, and (iv) positional bias visualization.
We introduce our SR Transformer using a $64\times64$ attention window, termed as \textbf{SST}~(12M), and a parameter-scaled-up variant \textbf{SST-L}~(20M).
Models using a larger attention window~(up to $96\times96$) are denoted with "$+$".
Since SR methods are typically trained on $64\times64$ cropped patches, naively applying a $96\times96$ window would introduce substantial padding to match the feature size to the window size, which can destabilize optimization.
Therefore, we train the "$+$" variants with $96\times96$ training patches, and this setting is marked with "$^{\dagger}$".
Despite the larger training patches, our method remains faster and more memory-efficient than prior methods trained with $64\times64$ patches, thanks to FlashAttention enabled by RIB, as shown in Section~\ref{sec:CostCompare}.
Implementation/testing details and additional experiments are provided in the Appendix.

\begin{table}[!t]
\centering
\caption{
    Comparison of training and inference costs across SR methods.
    Training costs are measured using the batch size of 10, which is the largest batch size that does not cause an OOM in all cases.
    Inference costs are measured while reconstructing a 1280$\times$720 image at scale $\times$2.
    All statistics are measured using an H200 GPU at FP32 precision.
    The best result is bolded.
    We leverage FlashAttention3~\cite{FlashAttn3} for the hardware-efficient acceleration kernel.
    For more comparisons across various GPUs and FlashAttention implementations, please refer to the appendix.
}
\label{tab:sr_cost_comparison}
{\renewcommand{\arraystretch}{0.8}
\resizebox{\textwidth}{!}{%
\begin{tabular}{lccccccc}
\toprule
\multirow{2}{*}{\textbf{Method}} &
\multicolumn{3}{c}{\textbf{Training}} &
\multicolumn{4}{c}{\textbf{Inference}} \\
\cmidrule(lr){2-4}\cmidrule(lr){5-8}
& patch size & sec/step  & memory &
\#FLOPs & \#params & latency & memory \\
\midrule
HAT~\cite{HAT} & $64\times64$ & 0.432 & 32.4GB & 5.81T & 20.6M & 709.9ms & 9070MB \\
ATD~\cite{ATD} & $\mathbf{96}\times\mathbf{96}$ & 1.291 & 132.5GB & 6.07T & 20.1M & 1266.8ms & 6239MB \\
PFT~\cite{PFT}          & $64\times64$ & 1.410 & 85.4GB & \textbf{5.03}T & \textbf{19.6}M & 1920.8ms & 28869MB \\
MambaIR~\cite{MambaIR}      & $64\times64$ & 0.791 & 50.0GB & 5.87T & 20.4M & 1043.8ms & 7704MB \\
MambaIRV2-L~\cite{MambaIRV2} & $64\times64$ & 1.210 & 89.8GB & 9.42T & 34.1M & 2308.8ms & 5816MB \\
\midrule
SST-L        & $64\times64$ & \textbf{0.370} & \textbf{31.4}GB & 18.8T & 20.3M & \textbf{608.4}ms & \textbf{2823}MB \\
SST-L$+$     & $\mathbf{96}\times\mathbf{96}$ & 0.672 & 67.7GB & 28.7T & 20.3M & 644.9ms & 2989MB \\
\bottomrule
\end{tabular}
}
}
\vspace{-0.4cm}
\end{table}


\begin{table}[!t]
  \centering
  \caption{Quantitative comparison (PSNR/SSIM) on benchmark datasets trained on DF2K datasets~\cite{DIV2K, DF2KDataset}. $^{\dagger}$ denotes the method is trained using $96\times 96$ patches. The best result is bolded.}
  \label{tab:sr_benchmark_df2k}

  {\renewcommand{\arraystretch}{0.75}
  \resizebox{\textwidth}{!}{%
    \begin{tabular}{@{} l c c c *{5}{c} @{}}
    \toprule
    \multirow{2}{*}{Method} &
    \multirow{2}{*}{Venue} &
    \multirow{2}{*}{scale} &
    \multirow{2}{*}{\#params} &
    \multicolumn{1}{c}{Set5} &
    \multicolumn{1}{c}{Set14} &
    \multicolumn{1}{c}{BSD100} &
    \multicolumn{1}{c}{Urban100} &
    \multicolumn{1}{c}{Manga109} \\
    \cmidrule(lr){5-9}
    & & & & PSNR/SSIM & PSNR/SSIM & PSNR/SSIM & PSNR/SSIM & PSNR/SSIM \\
    \midrule
    
    SwinIR~\cite{SwinIR}      & ICCVW`21    & \multirow{16}{*}{$\times 2$} & 11.8 & 38.42/0.9623 & 34.46/0.9250 & 32.53/0.9041 & 33.81/0.9427 & 39.92/0.9797 \\
    ESC$^{\dagger}$~\cite{ESC}     & ICCV`25     &                              & 12.5  &  38.59/0.9630 & 34.70/0.9259 & 32.61/0.9052 & 34.49/0.9466 & 40.38/0.9809 \\
    CAT-A~\cite{CAT}       & NeurIPS`22  &                              & 16.5  & 38.51/0.9626 & 34.78/0.9265 & 32.59/0.9047 & 34.26/0.9440 & 40.10/0.9805 \\
    ART~\cite{ART}         & ICLR`23     &                              & 16.4  & 38.56/0.9629 & 34.59/0.9267 & 32.58/0.9048 & 34.30/0.9452 & 40.24/0.9808 \\
    
    HAT~\cite{HAT}         & CVPR`23     &                              & 20.6  & 38.63/0.9630 & 34.86/0.9274 & 32.62/0.9053 & 34.45/0.9466 & 40.26/0.9809 \\
    ATD$^{\dagger}$~\cite{ATD}         & CVPR`24     &                              & 20.1  & 38.61/0.9629 & 34.95/0.9276 & 32.65/0.9056 & 34.70/0.9476 & 40.37/0.9810 \\
    PFT~\cite{PFT}         & CVPR`25     &                              & 19.6  & \textbf{38.68}/\textbf{0.9635} & \textbf{35.00}/\textbf{0.9280} & \textbf{32.67}/\textbf{0.9058} & 34.90/0.9490 & 40.49/\textbf{0.9815} \\
    
    \cmidrule(lr){1-2}\cmidrule(lr){4-9}
    
    IPG~\cite{IPG}      & CVPR`24  &                              & 20.4  & 38.61/0.9632 & 34.73/0.9270 & 32.60/0.9052 & 34.48/0.9464 & 40.24/0.9810 \\

    \cmidrule(lr){1-2}\cmidrule(lr){4-9}
    
    MambaIR~\cite{MambaIR}      & ECCV`24  &                              & 20.4  & 38.57/0.9627 & 34.67/0.9261 & 32.58/0.9048 & 34.15/0.9446 & 40.28/0.9806 \\
    MambaIRV2-S~\cite{MambaIRV2}  & CVPR`25  &                              & 9.6   & 38.53/0.9627 & 34.62/0.9256 & 32.59/0.9048 & 34.24/0.9454 & 40.27/0.9808 \\
    MambaIRV2-B~\cite{MambaIRV2}  & CVPR`25  &                              & 22.9  & 38.65/0.9631 & 34.89/0.9275 & 32.62/0.9053 & 34.49/0.9468 & 40.42/0.9810 \\
    MambaIRV2-L~\cite{MambaIRV2}  & CVPR`25  &                              & 34.1  & 38.65/0.9632 & 34.93/0.9276 & 32.62/0.9053 & 34.60/0.9475 & 40.55/0.9807 \\
    
    \cmidrule(lr){1-2}\cmidrule(lr){4-9}
    
    SST         & --         &                              & 11.7 & 38.63/0.9629 & 34.72/0.9258 & 32.55/0.9042 & 34.61/0.9474 & 40.31/0.9808 \\
    SST+$^{\dagger}$        & --         &                              & 11.7 & 38.63/0.9631 & 34.89/0.9273 & 32.62/0.9053 & 34.88/0.9488 & 40.44/0.9812 \\
    SST-L       & --         &                              & 20.3    & 38.65/0.9631 & 34.86/0.9270 & 32.60/0.9050 & 34.78/0.9485 & 40.44/0.9812 \\
    SST-L+$^{\dagger}$      & --         &                              & 20.3    & 38.64/0.9630 & 34.94/0.9273 & 32.61/0.9050 & \textbf{35.01}/\textbf{0.9497} & \textbf{40.56}/\textbf{0.9815} \\
    \midrule

    SwinIR~\cite{SwinIR}      & ICCVW`21    & \multirow{16}{*}{$\times 3$} & 11.9 & 34.97/0.9318 & 30.93/0.8534 & 29.46/0.8145 & 29.75/0.8826 & 35.12/0.9537 \\
    ESC$^{\dagger}$~\cite{ESC}         & ICCV`25     &                              & 12.5 & 35.14/0.9330 & 31.10/0.8552 & 29.53/0.8167 & 30.23/0.8895 & 35.60/0.9555 \\
    CAT-A~\cite{CAT}       & NeurIPS`22  &                              & 16.6  & 35.06/0.9326 & 31.04/0.8538 & 29.52/0.8160 & 30.12/0.8862 & 35.38/0.9546 \\
    ART~\cite{ART}         & ICLR`23     &                              & 16.6  & 35.07/0.9325 & 31.02/0.8541 & 29.51/0.8159 & 30.10/0.8871 & 35.39/0.9548 \\
    
    HAT~\cite{HAT}         & CVPR`23     &                              & 20.8  & 35.07/0.9329 & 31.08/0.8555 & 29.54/0.8167 & 30.23/0.8896 & 35.53/0.9552 \\
    ATD$^{\dagger}$~\cite{ATD}         & CVPR`24     &                              & 20.3  & 35.11/0.9330 & 31.13/0.8556 & 29.57/0.8176 & 30.46/0.8917 & 35.63/0.9558 \\
    PFT~\cite{PFT}         & CVPR`25     &                              & 19.8  & 35.15/0.9333 & 31.16/0.8561 & \textbf{29.58}/\textbf{0.8178} & 30.56/0.8931 & 35.67/0.9560 \\
    
    \cmidrule(lr){1-2}\cmidrule(lr){4-9}

    IPG~\cite{IPG}      & CVPR`24  &                              & 18.3  & 35.10/0.9332 & 31.10/0.8554 & 29.53/0.8168 & 30.36/0.8901 & 35.53/0.9554 \\

    \cmidrule(lr){1-2}\cmidrule(lr){4-9}
    
    MambaIR~\cite{MambaIR}      & ECCV`24  &                              & 20.4  & 35.08/0.9323 & 30.99/0.8536 & 29.51/0.8157 & 29.93/0.8841 & 35.43/0.9546 \\
    MambaIRV2-S~\cite{MambaIRV2}  & CVPR`25  &                              & 9.8   & 35.09/0.9326 & 31.07/0.8547 & 29.51/0.8157 & 30.08/0.8871 & 35.44/0.9549 \\
    MambaIRV2-B~\cite{MambaIRV2}  & CVPR`25  &                              & 23.1  & 35.18/0.9334 & 31.12/0.8557 & 29.55/0.8169 & 30.28/0.8905 & 35.61/0.9556 \\
    MambaIRV2-L~\cite{MambaIRV2}  & CVPR`25  &                              & 34.2  & 35.16/0.9334 & \textbf{31.18}/\textbf{0.8564} & 29.57/0.8175 & 30.34/0.8912 & 35.72/0.9561 \\
    
    \cmidrule(lr){1-2}\cmidrule(lr){4-9}
    
    SST         & --         &                              & 11.9 & 35.10/0.9323 & 31.00/0.8533 & 29.46/0.8149 & 30.35/0.8909 & 35.56/0.9554 \\
    SST+$^{\dagger}$        & --         &                              & 11.9 & 35.19/0.9335 & 31.08/0.8553 & 29.54/0.8170 & 30.59/0.8941 & 35.66/0.9560 \\
    SST-L       & --         &                              & 20.7    & 35.13/0.9328 & 31.02/0.8544 & 29.47/0.8152 & 30.52/0.8934 & 35.61/0.9557 \\
    SST-L+$^{\dagger}$      & --         &                              & 20.7    & \textbf{35.25}/\textbf{0.9337} & 31.16/0.8550 & 29.56/0.8174 & \textbf{30.75}/\textbf{0.8962} & \textbf{35.83}/\textbf{0.9568} \\
    \midrule
    
    SwinIR~\cite{SwinIR}      & ICCVW`21    & \multirow{16}{*}{$\times 4$} & 11.9 & 32.92/0.9044 & 29.09/0.7950 & 27.92/0.7489 & 27.45/0.8254 & 32.03/0.9260 \\
    ESC$^{\dagger}$~\cite{ESC}         & ICCV`25     &                              & 12.5 & 33.00/0.9054 & 29.21/0.7968 & 27.95/0.7504 & 27.89/0.8351 & 32.54/0.9295 \\
    CAT-A~\cite{CAT}       & NeurIPS`22  &                              & 16.6  & 33.08/0.9052 & 29.18/0.7960 & 27.99/0.7510 & 27.89/0.8339 & 32.39/0.9285 \\
    ART~\cite{ART}         & ICLR`23     &                              & 16.6  & 33.04/0.9051 & 29.16/0.7958 & 27.97/0.7510 & 27.77/0.8321 & 32.31/0.9283 \\
    
    HAT~\cite{HAT}         & CVPR`23     &                              & 20.8  & 33.04/0.9056 & 29.23/0.7973 & 28.00/0.7517 & 27.97/0.8368 & 32.48/0.9292 \\
    ATD$^{\dagger}$~\cite{ATD}         & CVPR`24     &                              & 20.3  & 33.10/0.9058 & 29.24/0.7974 & 28.01/0.7526 & 28.17/0.8404 & 32.62/0.9306 \\
    PFT~\cite{PFT}         & CVPR`25     &                              & 19.8  & 33.15/\textbf{0.9065} & \textbf{29.29}/0.7978 & \textbf{28.02}/\textbf{0.7527} & 28.20/0.8412 & 32.63/0.9306 \\
    
    \cmidrule(lr){1-2}\cmidrule(lr){4-9}

    IPG~\cite{IPG}      & CVPR`24  &                              & 17.0  & 33.15/0.9062 & 29.24/0.7973 & 27.99/0.7519 & 28.13/0.8392 & 32.53/0.9300 \\

    \cmidrule(lr){1-2}\cmidrule(lr){4-9}
    
    MambaIR~\cite{MambaIR}      & ECCV`24  &                              & 20.4  & 33.03/0.9046 & 29.20/0.7961 & 27.98/0.7503 & 27.68/0.8287 & 32.32/0.9272 \\
    MambaIRV2-S~\cite{MambaIRV2}  & CVPR`25  &                              & 9.8   & 32.99/0.9037 & 29.23/0.7965 & 27.97/0.7502 & 27.73/0.8307 & 32.33/0.9276 \\
    MambaIRV2-B~\cite{MambaIRV2}  & CVPR`25  &                              & 23.1  & 33.14/0.9057 & 29.23/0.7975 & 28.00/0.7511 & 27.89/0.8344 & 32.57/0.9295 \\
    MambaIRV2-L~\cite{MambaIRV2}  & CVPR`25  &                              & 34.2  & \textbf{33.19}/0.9062 & \textbf{29.29}/\textbf{0.7982} & 28.01/0.7521 & 28.07/0.8383 & 32.66/0.9304 \\
    
    \cmidrule(lr){1-2}\cmidrule(lr){4-9}
    
    SST         & --         &                              & 12.1 & 33.15/0.9059 & 29.21/0.7966 & 27.96/0.7505 & 28.08/0.8389 & 32.52/0.9295 \\
    SST+$^{\dagger}$        & --         &                              & 12.1 & 33.00/0.9054 & 29.21/0.7971 & 27.96/0.7511 & 28.23/0.8427 & 32.66/0.9308 \\
    SST-L       & --         &                              & 21.3    & 33.08/0.9056 & 29.25/0.7980 & 27.96/0.7511 & 28.23/0.8432 & 32.64/0.9305 \\
    SST-L+$^{\dagger}$      & --         &                              & 21.3    & 33.12/0.9057 & 29.23/0.7973 & 27.98/0.7512 & \textbf{28.39}/\textbf{0.8466} & \textbf{32.85}/\textbf{0.9322} \\
    \bottomrule
  \end{tabular}}
  }
  \vspace{-0.5cm}
\end{table}

\subsection{Comparisons of Training and Inference Costs}\label{sec:CostCompare}
We first compare the training and inference costs of our method against representative SR baselines~\cite{HAT, ATD, PFT, MambaIR, MambaIRV2} to highlight the efficiency gains enabled by FlashAttention.
As reported in Table~\ref{tab:sr_cost_comparison}, SST-L achieves the lowest training time and memory, inference latency, and memory, despite leveraging a much larger attention window.
This demonstrates that FlashAttention provides substantial end-to-end benefits.
Next, we compare SST-L$+$ with ATD under the same $96\times96$ training patch.
SST-L$+$ is 1.92$\times$ faster in training while using 1.9$\times$ less memory.
This result indicates that naively increasing the training patch without a hardware-efficient kernel can introduce significant overhead, reinforcing the importance of FlashAttention for scaling SR Transformers.
We also compare SST-L$+$ with PFT.
Although SST-L$+$ requires $5.7\times$ more FLOPs, it uses about $10\times$ less inference memory and achieves roughly $3\times$ lower latency.
We argue that this discrepancy stems from the fact that FLOPs reductions do not necessarily reflect practical efficiency gains~\cite{ShuffleNetV2, PConv, SHViT, PLKSR}.
Although PFT reduces FLOPs through top-$k$ sparsification, it still needs to materialize $N \times N$ and $N \times K$ matrices for each window (including shifted-window attention), such as $\mathbf{S}$, a previous layer's score matrix $\mathbf{S}_{prev}$, and a sparsification index $\mathbf{I}$.
As a result, PFT is primarily bottlenecked by memory traffic rather than computation, which leads to higher latency and increased memory consumption.
Finally, even when processing high-resolution feature maps~(e.g., $640\times360$ features for $1280\times720$ reconstruction at $\times2$), SST-L$+$ remains more efficient in inference than Mamba-based~(linear-operator) SR methods, underscoring the practicality and promise of FlashAttention-equipped Transformers.

\subsection{Quantitative Results}
Next, we conduct a quantitative comparison to demonstrate the strong performance of our method. 

\subsubsection{Results on DF2K Datasets}
We compare our proposed networks~(SST, SST$+$, SST-L, and SST-L$+$), trained on the DF2K dataset~\cite{DIV2K, DF2KDataset}, with representative SR methods employing different core operators, including Transformers~\cite{SwinIR, ESC, CAT, ART, HAT, ATD, PFT}, a graph neural network~\cite{IPG}, and Mamba~\cite{MambaIR, MambaIRV2}.
All evaluations are conducted on commonly used benchmarks, including Set5~\cite{Set5}, Set14~\cite{Set14}, BSD100~\cite{BSD100}, Urban100~\cite{Urban100}, and Manga109~\cite{Manga109}.
As shown in Table~\ref{tab:sr_benchmark_df2k}, SST achieves superior performance on Urban100$\times$2, outperforming HAT~(20M) by 0.16\,dB with only 12M parameters, demonstrating the effectiveness of $64\times64$ attention.
These results demonstrate that substantial performance gains can be achieved simply by leveraging RIB to overcome the window-size limitation, which is commonly focused by existing SR methods, without scaling on training-patch or data.
Furthermore, when trained with $96\times96$ patches, SST$+$~(12M) surpasses ATD~(20M), trained under the same patch size, by 0.18\,dB on Urban100$\times$2, indicating that SST$+$ more effectively exploits the benefits of the large training patch.
Finally, SST-L and SST-L$+$ achieve strong performance on Urban100 and Manga109 across all upscaling factors.
Notably, these improvements come with substantially reduced training and inference costs, as demonstrated in Section~\ref{sec:CostCompare}.

\begin{table*}[t]
  \centering
  \caption{Quantitative comparison (PSNR/SSIM) on benchmark datasets trained on DFLIP datasets~\cite{DIV2K, DF2KDataset, LSDIR, DiverSeg}. $^{\dagger}$ denotes the method is trained using $96\times 96$ patches. The best result is bolded.}
  \label{tab:sr_benchmark_DFLIP}
  \vspace{-0.1cm}
  {\renewcommand{\arraystretch}{0.85}
  \resizebox{\textwidth}{!}{%
    \begin{tabular}{@{} l c c *{7}{c} @{}}
    \toprule
    \multirow{2}{*}{Method} &
    \multirow{2}{*}{scale} &
    \multirow{2}{*}{\#params} &
    \multicolumn{1}{c}{Set5} &
    \multicolumn{1}{c}{Set14} &
    \multicolumn{1}{c}{BSD100} &
    \multicolumn{1}{c}{Urban100} &
    \multicolumn{1}{c}{Manga109} &
    \multicolumn{1}{c}{DIV2K-val} &
    \multicolumn{1}{c}{LSDIR-val} \\
    \cmidrule(lr){4-10}
    & & & PSNR/SSIM & PSNR/SSIM & PSNR/SSIM & PSNR/SSIM & PSNR/SSIM & PSNR/SSIM & PSNR/SSIM \\
    \midrule

    PFT~\cite{PFT}                 & \multirow{4}{*}{$\times 2$} & 19.6 & 38.67/0.9633 & 35.11/0.9282 & 32.71/0.9062 & 35.24/0.9508 & 40.64/0.9817 & 37.17/0.9526 & 32.92/0.9321 \\
    MambaIRV2-L~\cite{MambaIRV2}   &                             & 34.2 & 38.77/0.9636 & 35.20/0.9287 & \textbf{32.74}/\textbf{0.9065} & 35.20/0.9508 & 40.82/0.9820 & 37.16/0.9524 & 32.93/0.9321 \\
    SST-L                         &                             & 20.3 & 38.70/0.9635 & 35.06/0.9274 & 32.69/0.9058 & 35.35/0.9515 & 40.64/0.9817 & 37.21/0.9528 & 32.97/0.9325 \\
    SST-L+$^{\dagger}$            &                             & 20.3 & \textbf{38.78}/\textbf{0.9637} & \textbf{35.28}/\textbf{0.9288} & 32.73/0.9064 & \textbf{35.63}/\textbf{0.9530} & \textbf{40.86}/\textbf{0.9821} & \textbf{37.32}/\textbf{0.9535} & \textbf{33.12}/\textbf{0.9338}  \\
    \midrule

    PFT~\cite{PFT}                 & \multirow{4}{*}{$\times 3$} & 19.8 & 35.12/0.9335 & 31.45/0.8572 & 29.60/0.8188 & 31.08/0.8996 & 35.81/0.9567 & 33.41/0.9020 & 29.16/0.8578 \\
    MambaIRV2-L~\cite{MambaIRV2}   &                             & 34.2 & 35.23/0.9341 & \textbf{31.56}/\textbf{0.8586} & \textbf{29.66}/\textbf{0.8195} & 31.13/0.9000 & 35.98/0.9574 & 33.43/0.9020 & 29.23/0.8586 \\
    SST-L                         &                             & 20.7 & 35.24/0.9339 & 31.47/0.8567 & 29.57/0.8177 & 31.22/0.9010 & 35.90/0.9570 & 33.45/0.9026 & 29.21/0.8587 \\
    SST-L+$^{\dagger}$            &                             & 20.7 & \textbf{35.33}/\textbf{0.9345} & 31.54/0.8576 & 29.64/0.8193 & \textbf{31.53}/\textbf{0.9046} & \textbf{36.12}/\textbf{0.9580} & \textbf{33.55}/\textbf{0.9037} & \textbf{29.33}/\textbf{0.8609} \\
    \midrule

    PFT~\cite{PFT}                 & \multirow{4}{*}{$\times 4$} & 19.8 & 33.17/0.9071 & 29.50/0.8005 & 28.03/0.7537 & 28.71/0.8515 & 32.84/0.9323 & 31.38/0.8573 & 27.22/0.7950 \\
    MambaIRV2-L~\cite{MambaIRV2}   &                             & 34.2 & 33.20/0.9073 & 29.57/\textbf{0.8018} & \textbf{28.11}/\textbf{0.7556} & 28.75/0.8526 & 32.99/0.9334 & 31.40/0.8576  & 27.29/0.7967 \\
    SST-L                         &                             & 21.3 & \textbf{33.25}/\textbf{0.9080}           & \textbf{29.58}/0.8016           & 28.07/0.7541           & 28.90/0.8549           & 33.00/0.9334           & 31.44/0.8584 & 27.31/0.7972 \\
    SST-L+$^{\dagger}$            &                             & 21.3 & 33.10/0.9067           & \textbf{29.58}/0.8016           & 28.06/0.7541           & \textbf{29.06}/\textbf{0.8583}           & \textbf{33.18}/\textbf{0.9345}           & \textbf{31.50}/\textbf{0.8594} & \textbf{27.37}/\textbf{0.7991} \\
    \bottomrule
    \end{tabular}}
  }
  \vspace{-0.4cm}
\end{table*}

\subsubsection{Results on DFLIP Datasets}
Our method exhibits remarkable gains in the 12M regime, even surpassing prior 20M parameter methods. 
However, when scaling to the 20M parameter regime, our results become comparable to existing approaches.
We hypothesize that this behavior stems from data limitations in standard SR training: given the strong representational capacity of our network, the commonly used training data may be insufficient, leading to overfitting.
To examine this hypothesis, we train our models as well as representative SOTA baselines~(\cite{PFT, MambaIRV2}) on the DFLIP training datasets, which augment the widely used DF2K dataset with additional large-scale datasets, such as LSDIR~\cite{LSDIR} and DiverSeg-IP~\cite{DiverSeg}.
We adopt the same training setup as in the DF2K experiments to ensure a fair comparison. 
In addition, to broaden the evaluation scope, we report results on DIV2K-val~\cite{DIV2K} and LSDIR-val~\cite{LSDIR}, which have been less commonly included in prior works.
As shown in Table~\ref{tab:sr_benchmark_DFLIP}, when trained on DF2K only, SST-L underperforms PFT by 0.12\,dB on Urban100$\times$2, whereas when trained on DFLIP, it surpasses PFT by 0.11\,dB, highlighting SST's strong data-scaling capability.
Moreover, SST-L$+$ consistently delivers substantial improvements on Urban100, Manga109, DIV2K-val, and LSDIR-val, demonstrating that our scaling design choices, including window size, training patch size, and data size, are highly effective.
Still, SST-L$+$ does not show consistent improvements on Set5, Set14, and BSD100 under the $\times$4 setting. 
We provide further analysis in Appendix Section~\ref{sec:boundary_analysis}.

\begin{table}[!t]
\centering
\caption{
    Training/inference costs and performance of attention/bias variants on SST$+$.
    Training and inference costs are measured as detailed in Table~\ref{tab:sr_cost_comparison}.
}
\label{tab:AblationAttention}
{\renewcommand{\arraystretch}{0.95}
\resizebox{\textwidth}{!}{
\begin{tabular}{ccccccccc}
\toprule
\multirow{2}{*}{\textbf{Attention}} &
\multirow{2}{*}{\textbf{Bias}} &
\multicolumn{2}{c}{\textbf{Training}} &
\multicolumn{3}{c}{\textbf{Inference}} &
\multicolumn{1}{c}{\textbf{Urban100}} &
\multicolumn{1}{c}{\textbf{Manga109}} \\
\cmidrule(lr){3-4}\cmidrule(lr){5-7}\cmidrule(lr){8-8}\cmidrule(lr){9-9}
& & sec/step & memory &
\#params & latency & memory & PSNR/SSIM & PSNR/SSIM \\
\midrule
Naive & RPB~\cite{SwinTransformer} & - & - & 13.6M & 11705.3ms & 126181MB & - / - & - / - \\
FlexAttention~\cite{FlexAttn, ESC} & RPB~\cite{SwinTransformer} & 5.048 & 47.2GB & 13.6M & 2414.0ms & 2487MB & 34.91/0.9489 & 40.42/0.9812 \\
FlashAttention3~\cite{FlashAttn3} & CPE~\cite{CPE} & 0.625 & 46.5GB & 11.3M & 537.3ms & 2354MB & \multicolumn{2}{c}{Not Converged} \\
FlashAttention3~\cite{FlashAttn3} & FlashBias~\cite{FlashBias} & 0.473 & 47.3GB & 46.7M & 455.7ms & 2825MB &  \multicolumn{2}{c}{Not Converged}  \\
FlashAttention3~\cite{FlashAttn3} & RoPE-ViT~\cite{RoPEViT} & 0.489 & 47.9GB & 11.3M & 432.7ms & 2915MB & 34.71/0.9479 & 40.40/0.9811 \\
\midrule
FlashAttention3~\cite{FlashAttn3} & RIB~(Ours) & 0.479 & 47.1GB & 11.7M & 455.7ms & 2825MB & 34.88/0.9488 & 40.44/0.9812 \\
\bottomrule
\end{tabular}
}
\vspace{-0.5cm}
}
\end{table}
\begin{table}[t]
\centering
\caption{Ablation studies of proposed components in the SST.}
\label{tab:ablation_parallel_psnrssim}
\renewcommand{\arraystretch}{0.9} 
\vspace{-0.15cm}
\begin{subtable}[t]{0.34\textwidth}
\centering
\caption{Gating Type}
\resizebox{\linewidth}{!}{%
\begin{tabular}{@{}lcc@{}}
\toprule
\textbf{Gating Type} & \textbf{Urban100} & \textbf{Manga109} \\
\midrule
Without $\mathbf{G}$ & \multicolumn{2}{c}{Not Converged} \\
$\mathrm{PWConv}$-only~\cite{GatedAttn} & 34.55/0.9473 & 40.24/0.9807 \\
\midrule
CLA & 34.61/0.9474 & 40.31/0.9808 \\
\bottomrule
\end{tabular}%
}
\end{subtable}
\hfill
\begin{subtable}[t]{0.33\textwidth}
\centering
\caption{Window Strategy}
\resizebox{\linewidth}{!}{%
\begin{tabular}{@{}lcc@{}}
\toprule
\textbf{Window Strategy} & \textbf{Urban100} & \textbf{Manga109} \\
\midrule
\{4, 8, 16, 32, 64, 64\} & 34.56/0.9470 & 40.22/0.9807 \\
\{64, 64, 32, 16, 8, 4\} & 34.51/0.9468 & 40.17/0.9804 \\
\{64, 64, 64, 64, 64, 64\} & 34.54/0.9468 & 40.19/0.9806 \\
\midrule
\{16, 32, 64, 16, 32, 64\} & 34.61/0.9474 & 40.31/0.9808 \\
\bottomrule
\end{tabular}%
}
\end{subtable}
\hfill
\begin{subtable}[t]{0.28\textwidth}
\centering
\caption{Fourier Embedding}
\resizebox{\linewidth}{!}{%
\begin{tabular}{@{}lcc@{}}
\toprule
\textbf{Fourier Emb.} & \textbf{Urban100} & \textbf{Manga109} \\
\midrule
$L$=5 & 34.20/0.9447 & 39.96/0.9797 \\
$L$=15 & 34.48/0.9468 & 40.17/0.9805 \\
SIREN~\cite{SIREN} & \multicolumn{2}{c}{Not Converged} \\
\midrule
$L$=10 & 34.61/0.9474 & 40.31/0.9808 \\
\bottomrule
\end{tabular}%
}
\end{subtable}
\vspace{-0.3cm}
\end{table}

\subsection{Ablation Studies}
Next, we conduct ablation studies to verify that our proposal is effective.
As shown in Table~\ref{tab:AblationAttention}, our proposed RIB achieves almost the same performance as FlexAttention with RPB~\cite{FlexAttn, ESC}, while offering substantially improved efficiency~(e.g., \emph{10.5$\times$ faster training and 5$\times$ faster inference}).
This suggests that RIB preserves the effectiveness of RPB despite its factorized approximation.
Next, we compare RIB with methods that cannot directly provide a spatial prior to $\mathbf{S}$.
Although FlashBias~\cite{FlashBias} introduces attention bias in a similar manner to RIB, it is designed for post-training acceleration by decomposing learned biases into low-rank positional embeddings.
Thus, when trained from scratch, its decomposed positional embeddings are learned from zero initialization and fail to provide an effective spatial prior to $\mathbf{S}$, resulting in divergence.
Similarly, CPE~\cite{CPE} also diverges, as it promotes locality only in the feature map rather than directly injecting a spatial prior into $\mathbf{S}$.
Finally, RIB demonstrates superior performance over RoPE-ViT~\cite{RoPEViT}, which is proposed to enhance 2D RoPE for ViTs, suggesting that preserving pixel integrity is effective.

In addition, Table~\ref{tab:ablation_parallel_psnrssim} presents further ablations on our proposed CLA, the cyclic window strategy, and the number of frequency bands used in RIB~($L$).
First, when the $\mathbf{G}$ path is entirely removed, training diverges, while this issue is alleviated by incorporating a $\mathrm{PWConv}$-only $G$.
We interpret this behavior as the gating path mitigating the training instability induced by the high entropy of the large window and also by the high learning rate, which is partially consistent with findings from prior work~\cite{GatedAttn}.
Please note that RPB with FlexAttention is also diverged when trained without $\mathbf{G}$.
Introducing CLA with an additional $\mathrm{DWConv}$ yields further performance improvements, demonstrating its effectiveness.
Next, the cyclic window strategy outperforms monotonically increasing/decreasing schedules and a fixed $64\times64$ window, suggesting that its repeated local-to-broad pattern balances multi-scale feature extraction with a scan-and-focus behavior and provides powerful performance.
Finally, setting $L=10$ delivers the best performance compared to other numbers or using SIREN~\cite{SIREN} activations. 
This observation is consistent with findings reported in previous research~\cite{NERF, LHF} on the novel view synthesis task.

\begin{figure}[t]
  \centering
  \includegraphics[width=0.9\textwidth]{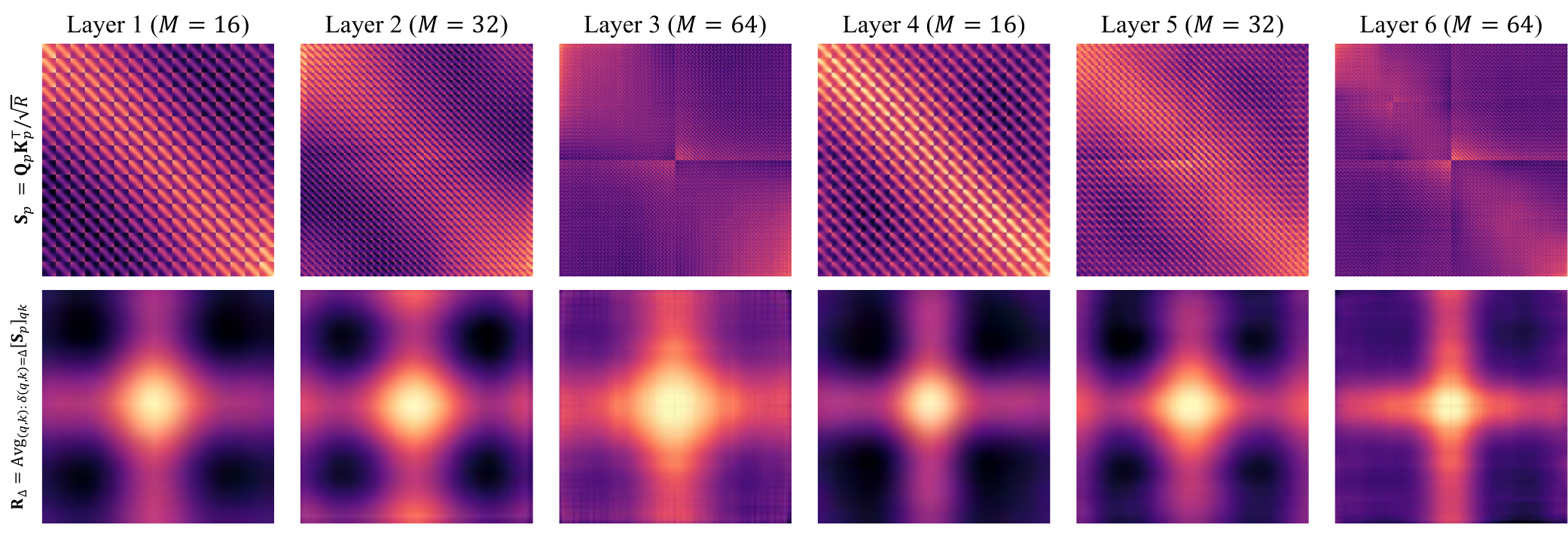}
  \vspace{-0.05cm}
  \caption{
    Visualized positional score matrix~($\mathbf{S}_{\mathrm{p}}$) and relative positional bias table~($\mathbf{R}_{\Delta}$) calculated from the second block of SST. 
    We average the biases that correspond to the same 2D distance, since positional bias predicted by our RIB does not guarantee equivalence across the same relative offsets.
  }
  \label{fig:PB}
  \vspace{-0.5cm}
\end{figure}

\vspace{-0.1cm}
\subsection{Positional Bias Visualization}
\vspace{-0.1cm}
To verify whether RIB captures meaningful positional bias, we conduct a visualization analysis.
As shown in Figure~\ref{fig:PB}, RIB assigns high correlations to nearby pixels while also inducing strong vertical and horizontal correlations.
This indicates that RIB provides a spatial prior for self-attention that is well suited to image structures.
Moreover, the visualization aligns with the intended behavior of the cyclic window strategy.
The first cycle exhibits a broader bias pattern, suggesting a coarse scan over wider spatial contexts, whereas the second cycle shows a more concentrated bias, implying refined and focused aggregation.
This supports our design, where the early cycle captures broad contextual information, and the later cycle performs more localized refinement.
\vspace{-0.1cm}
\section{Conclusion}
\vspace{-0.1cm}
In this paper, we identify conventional RPB as an efficiency bottleneck that prevents SR Transformers from scaling with hardware-efficient attention. 
To overcome this limitation, we propose RIB, a FlashAttention-compatible attention bias that retains the representational power of RPB while preserving the integrity of pixel representations.
Together with CLA and a cyclic window strategy, RIB enables SST to scale attention window size, training patch size, and dataset size more efficiently. 
As a result, SST substantially reduces training and inference costs while achieving 29.06\,dB PSNR on Urban100$\times$4, suggesting a promising yet underexplored direction for scaling SR Transformers through hardware-efficient attention design.

\bibliographystyle{plainnat}
\bibliography{main}

\clearpage
\appendix
The appendix provides further supporting analysis in the main manuscript.
We first summarize related work on CNN-based SR methods, SR Transformers, alternative sub-quadratic operators, and hardware-efficient acceleration methods.
We then provide implementation, training, and testing details of the proposed SST.
In addition, we present extended experiments on parameter scaling, comparisons of training and inference costs across FlashAttention implementations, and inference-cost comparisons across different devices.
We also report results on lightweight SR tasks and investigate the applicability of the proposed RIB to hybrid CNN-Transformer architectures.
Finally, we include additional visual quality comparisons, Local Attribution Map~(LAM) analyses, qualitative analyses of CLA and PWConv-only gating, and discuss the limitations and future directions of our methods.

\section{Related Work}
\subsection{Traditional CNNs}
Early deep learning-based SR methods predominantly relied on a convolution operation, which is well suited for extracting local features from the input~\cite{SRCNN, FSRCNN, ESPCN, EDSR, RCAN, RCANIT}.
Motivated by prior findings that stacking small convolutions can gradually expand the effective receptive field while keeping the parameter count relatively modest~\cite{VGGNet}, many convolution-based works adopted designs that heavily stack 3$\times$3 convolutions.
However, such designs only model long-range dependencies indirectly, which limits their ability to capture similar but spatially distant patterns~\cite{SwinIR}.
Moreover, these stacked 3$\times$3 convolutional structures often introduce large parameter counts, frequently leading to over-parameterization~\cite{PFS}.

\subsection{Transformers}
To address these limitations, Transformers~\cite{Transformers}, which leverage self-attention as a core operator, have garnered significant attention in SR tasks.
By performing input-adaptive global feature aggregation, self-attention offers strong representational capacity and can potentially improve performance with fewer parameters and lower computational cost.
Nevertheless, unlike many vision tasks that treat an image patch~(e.g., 16$\times$16) as a token, SR commonly operates at the pixel level and processes high-resolution inputs, making it difficult to apply vanilla self-attention directly due to its quadratic complexity.
Therefore, many SR Transformers restrict self-attention to local windows to address the quadratic complexity of self-attention~\cite{SwinTransformer}.
While this makes the computational cost more practical and manageable, it also limits long-range modeling capability.
As a result, much recent effort has focused on expanding the receptive field while reducing the number of pixels involved in self-attention computation~\cite{SwinIR, SRFormer, DAT, ART, CAT, HAT, CFAT}.

\subsection{Alternative Operators}
Another actively studied direction is to replace self-attention with sub-quadratic operators.
Representative examples include proposing attention-like operators with sub-quadratic complexity~\cite{Restormer, RGT, NGSwin, ATD}, adapting Mamba~\cite{Mamba}, which has attracted substantial interest in sequence/language modeling, to the SR domain~\cite{MambaIR, MambaIRV2, TSPMamba}, and integrating sparsification into self-attention~\cite{PFT, IET}.
For lightweight SR, where efficiency under small model size is a primary goal, leveraging large-kernel convolutions~\cite{ShuffleMixer, SMFANet, PLKSR, LKMN} and reducing the number of attention-map computations~\cite{ELAN, ASID, UPS} have also been considered.

\begin{figure}[t]
  \centering
  \includegraphics[width=\textwidth]{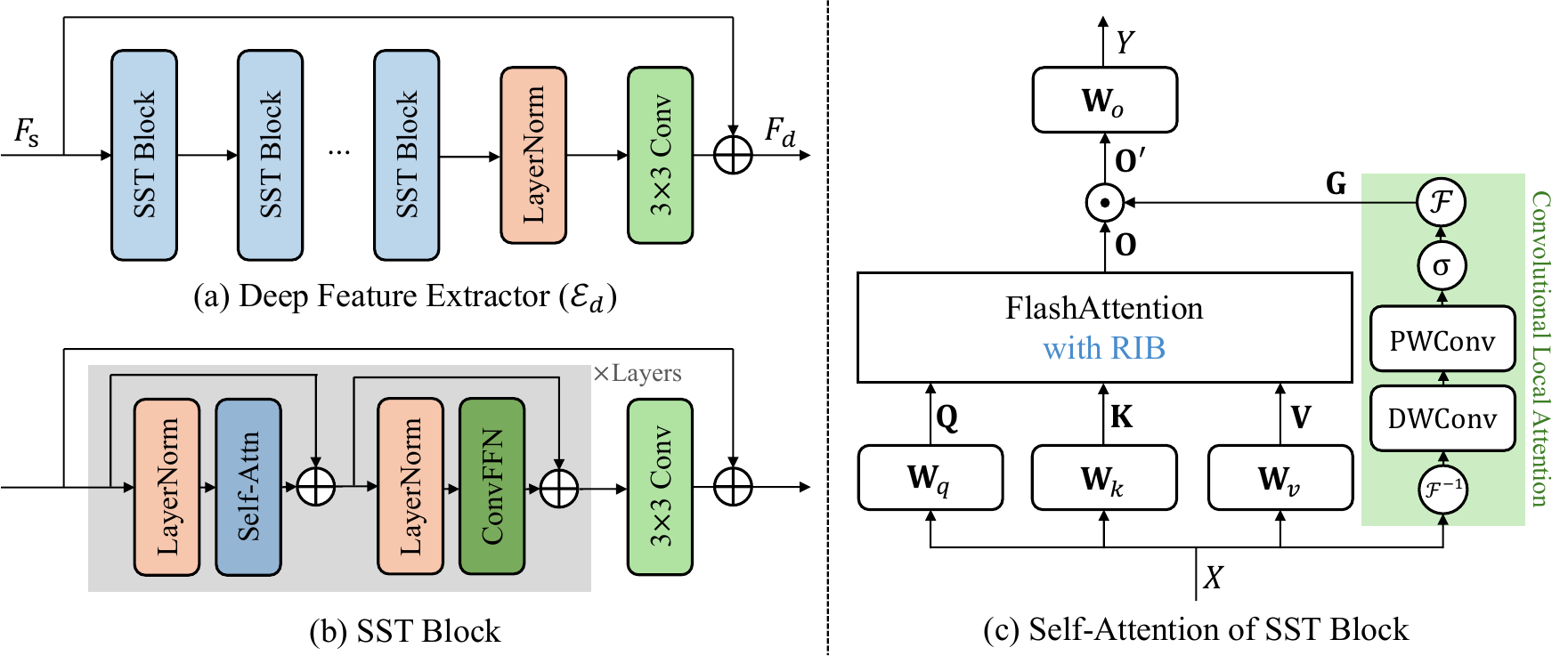}
  \caption{
    Overall illustration for SST architecture.
  }
  \label{fig:RINImplementation}
\end{figure}

\subsection{Hardware-Efficient Acceleration Methods}
However, most of the aforementioned approaches focus mainly on computational complexity and comparatively overlook memory efficiency. 
In practice, when data movement becomes the bottleneck, cutting floating-point operations~(FLOPs) alone often fails to reduce latency and memory usage.
For instance, self-attention repeatedly moves large $N\times N$ matrices~(e.g., $\mathbf{S}$ and $\mathbf{P}$) between HBM and SRAM during execution, which can incur substantial latency and memory overhead that FLOPs fail to account for.
To directly mitigate this issue, FlashAttention~\cite{FlashAttn, FlashAttn2, FlashAttn3} was introduced to reduce memory traffic while still computing exact attention, thereby significantly lowering latency and memory usage.
Due to its practical impact, FlashAttention has been rapidly adopted across diverse Transformer applications, and is becoming a key ingredient for scaling Transformers in terms of both model size and sequence length.
In SR, however, the relative position bias~(RPB) that is crucial for performance~\cite{T5, SwinTransformer} is not compatible with the layouts supported by FlashAttention, making it difficult to leverage FlashAttention directly.
Several works have attempted to resolve this limitation~\cite{FlashBias, ESC}. 
Among these works, FlashBias~\cite{FlashBias} is similar to our method in that it injects positional priors via channel-wise concatenation. 
However, since it was proposed for post-training acceleration, it has the drawback that, when trained from scratch, embeddings with the same shape cannot effectively provide positional priors to self-attention, making training difficult.
In addition, the acceleration in ESC~\cite{ESC} based on FlexAttention~\cite{FlexAttn} is substantially slower during training than FlashAttention kernels. 
Moreover, it is tied to specific frameworks such as PyTorch and increases system complexity due to its reliance on low-level compilation.
To train MambaIRV2-L~\cite{MambaIRV2} and PFT~\cite{PFT} on DFLIP datasets, we followed the training configurations reported in their respective papers and official codebases.

\begin{table}[t]
\centering
\caption{Network implementation details.}
\label{tab:network_implementation}
\resizebox{\textwidth}{!}{%
\begin{tabular}{lccccccccc}
\toprule
Methods & $D$ & Blocks & Layers & Window Sizes & Heads & $L$ & $d_h$ & $R$ & ConvFFN Exp. \\
\midrule
SST-light & 48 & 5 & 6 & [8, 16, 32, 16, 32, 64] & 3 & 10 & 32 & [16, 16, 16, 24, 24, 24] & 1.5 \\
SST-light+ & 48 & 5 & 6 & [16, 32, 48, 32, 48, 96] & 3 & 10 & 32 & [16, 16, 16, 24, 24, 24] & 1.5 \\
\midrule
SST & 180 & 6 & 6 & [16, 32, 64, 16, 32, 64] & 6 & 10 & 32 & [18, 18, 18, 34, 34, 34] & 1.25 \\
SST+ & 180 & 6 & 6 & [16, 32, 48, 32, 48, 96] & 6 & 10 & 32 & [18, 18, 18, 34, 34, 34] & 1.25 \\
\midrule
SST-L & 192 & 8 & 6 & [16, 32, 64, 16, 32, 64] & 6 & 10 & 32 & [16, 16, 16, 32, 32, 32] & 2 \\
SST-L+ & 192 & 8 & 6 & [16, 32, 48, 32, 48, 96] & 6 & 10 & 32 & [16, 16, 16, 32, 32, 32] & 2 \\
\midrule
SST-XL+ & 224 & 10 & 6 & [16, 32, 48, 32, 48, 96] & 7 & 10 & 32 & [16, 16, 16, 32, 32, 32] & 2 \\
\bottomrule
\end{tabular}
}
\end{table}
\begin{table}[t]
  \centering
  \caption{
      Training configuration across our networks for $\times$2 upscaling. 
      For $\times$3 and $\times$4, we initialize from the weights of $\times$2 and $\times$3, respectively~\cite{IGConv}, and halve the number of iterations, learning rate, and schedule.
  }
  \label{tab:train-config}
  \resizebox{\textwidth}{!}{%
  \begin{tabular}{lccccccccc}
    \toprule
    Methods & TrainingPatch & Dataset & BatchSize & Iteration & Optimizer & LR & Schedule~($\gamma=0.5$) & Loss \\
    \midrule
    SST-light  & 64$\times$64 & DIV2K & \multirow{2}{*}{64} & \multirow{7}{*}{500000} & \multirow{7}{*}{AdamW} & \multirow{7}{*}{5e-4} & \multirow{7}{*}{\shortstack{[250000, 400000,\\ 450000, 475000, 490000]}} & \multirow{7}{*}{L1Loss} \\
    SST-light$+$ & 96$\times$96 & or DFLIP &  &  &  &  &  &  \\
    \cmidrule(lr){1-4}
    SST  & 64$\times$64 & \multirow{2}{*}{DF2K} & \multirow{5}{*}{32} &  &  &  &  &  \\
    SST$+$ & 96$\times$96 &  &  &  &  &  &  &  \\
    \cmidrule(lr){1-3}
    SST-L  & 64$\times$64 & DF2K  &  &  &  &  &  &  \\
    SST-L$+$ & 96$\times$96 & or DFLIP  &  &  &  &  &  &  \\
    \cmidrule(lr){1-3}
    SST-XL$+$ & 96$\times$96 & DFLIP  &  &  &  &  &  &  \\
    \bottomrule
  \end{tabular}
  }
\end{table}

\section{Implementation, Training and Testing Details}
Our network follows the same overall architecture as the standard SR Transformer~\cite{SwinIR}, and its detailed configuration is provided in Figure~\ref{fig:RINImplementation} and Table~\ref{tab:network_implementation}.
On top of the window-based self-attention layer~\cite{SwinIR}, we incorporate only two proposed components: the proposed Rank-factorized Implicit Neural Bias~(RIB) and a convolutional local-attention module~(CLA).
We also adopt the ConvFFN introduced in prior work~\cite{SRFormer}, setting the kernel size to 3 to strengthen high-frequency feature extraction.
Inspired by previous work~\cite{HiTSR}, we replace shifted windows with a cyclic window strategy, which periodically increases the window sizes.
This design aims to extract multi-scale features while enabling more aggressive use of large windows.
Among the RIB hyper-parameters $(L, d_{h}, R)$, we set $L=10$ following the ablation results in Table~\textcolor{red}{5} in the main manuscript, and choose $d_{h}$ and $R$ empirically.
In particular, for $R$, we select a value such that the dimensionality $(D+R)$, after combining positional tokens with pixel tokens, is a multiple of 8.
This allows us to match the FlashAttention layout without channel-wise zero padding.
Moreover, in line with the main objective of the cyclic window strategy~(scan-and-focus), we use a larger value of $R$ in the second cycle~(e.g., $\{18, 18, 18, 34, 34, 34\}$).
This setting achieves PSNR gains of 0.46\,dB and 0.07\,dB on Urban100$\times$2 compared to using the $R$ in the reverse order~(e.g., $\{34, 34, 34, 18, 18, 18\}$) and using the same $R$ throughout all cycles~(e.g., $\{34, 34, 34, 34, 34, 34\}$), respectively.
For $d_h$, setting it to 16 leads to a 0.21\,dB performance drop on Urban100$\times$2, whereas setting it to 64 yields almost no difference compared to 32.
Therefore, we set $d_h$ to 32 in our experiments.
For our experiments, we use PyTorch version 2.9.1 and implement our method using the widely used BasicSR toolbox~\cite{BasicSR}.

As shown in the training configuration in Table~\ref{tab:train-config}, our overall training setup is largely consistent with prior studies.
However, to highlight our main focus on scalability, we additionally consider variants trained with larger patch sizes~(e.g., $96\times96$) and variants trained on larger-scale datasets~(e.g., DFLIP~\cite{DIV2K, DF2KDataset, LSDIR, DiverSeg}).

For quantitative comparison, we crop the boundary of the restored image $I_{SR}$ by the upscaling factor, convert the result to the YCbCr color space, and then compute PSNR and SSIM on the Y channel, following standard convention.
Please note that we compare against methods for which both the official code and pretrained weights are publicly available.

\begin{table*}[t]
  \centering
  \caption{Boundary crop sensitivity comparison on benchmark datasets trained on DFLIP datasets~\cite{DIV2K, DF2KDataset, LSDIR, DiverSeg}. $^{\dagger}$ denotes the method is trained using $96\times 96$ patches. The best result is bolded.}
  \label{tab:crop_compare}
  {\renewcommand{\arraystretch}{1}
  \resizebox{\textwidth}{!}{%
    \begin{tabular}{@{} l c c *{7}{c} @{}}
    \toprule
    \multirow{2}{*}{Method} &
    \multirow{2}{*}{scale} &
    \multirow{2}{*}{\#params} &
    \multicolumn{1}{c}{Set5} &
    \multicolumn{1}{c}{Set14} &
    \multicolumn{1}{c}{BSD100} &
    \multicolumn{1}{c}{Urban100} &
    \multicolumn{1}{c}{Manga109} &
    \multicolumn{1}{c}{DIV2K-val} &
    \multicolumn{1}{c}{LSDIR-val} \\
    \cmidrule(lr){4-10}
    & & & PSNR/SSIM & PSNR/SSIM & PSNR/SSIM & PSNR/SSIM & PSNR/SSIM & PSNR/SSIM & PSNR/SSIM \\
    \midrule

    PFT~\cite{PFT}
    & \multirow{8}{*}{$\times 4$}
    & \multirow{3}{*}{19.8}
    & 33.17/0.9071
    & 29.50/0.8005
    & 28.03/0.7537
    & 28.71/0.8515
    & 32.84/0.9323
    & 31.38/0.8573
    & 27.22/0.7950 \\
    \quad\(\hookrightarrow\) w/o crop
    & & & 32.85/0.9066 & 28.90/0.8009 & 27.90/0.7542 & 28.61/0.8515 & 32.80/0.9324 & 31.36/0.8574 & 27.19/0.7953 \\
    \quad\(\hookrightarrow\) \(\Delta\)
    & & & \blue{-0.32}/\blue{-0.0005}
    & \blue{-0.60}/\red{+0.0004}
    & \blue{-0.13}/\red{+0.0005}
    & \blue{-0.10}/+0.0000
    & \blue{-0.04}/\red{+0.0001}
    & \blue{-0.02}/\red{+0.0001}
    & \blue{-0.03}/\red{+0.0003} \\
    \addlinespace[0.15em]

    MambaIRV2-L~\cite{MambaIRV2}
    & 
    & \multirow{3}{*}{34.2}
    & 33.20/0.9073
    & 29.57/0.8018
    & 28.11/0.7556
    & 28.75/0.8526
    & 32.99/0.9334
    & 31.40/0.8576
    & 27.29/0.7967 \\
    \quad\(\hookrightarrow\) w/o crop
    & & & 32.89/0.9070 & 28.94/0.8021 & 28.06/0.7563 & 28.67/0.8527 & 32.96/0.9336 & 31.39/0.8578 & 27.26/0.7970 \\
    \quad\(\hookrightarrow\) \(\Delta\)
    & & & \blue{-0.31}/\blue{-0.0003}
    & \blue{-0.63}/\red{+0.0003}
    & \blue{-0.05}/\red{+0.0007}
    & \blue{-0.08}/\red{+0.0001}
    & \blue{-0.03}/\red{+0.0002}
    & \blue{-0.01}/\red{+0.0002}
    & \blue{-0.03}/\red{+0.0003} \\
    \addlinespace[0.15em]

    SST-L+$^{\dagger}$
    &
    & \multirow{3}{*}{21.3}
    & 33.10/0.9067
    & 29.58/0.8016
    & 28.06/0.7541
    & 29.06/0.8583
    & 33.18/0.9345
    & 31.50/0.8594
    & 27.37/0.7991 \\
    \quad\(\hookrightarrow\) w/o crop
    & & & 32.97/0.9063 & 29.36/0.8022 & 28.08/0.7548 & 29.02/0.8584 & 33.22/0.9347 & 31.51/0.8596 & 27.38/0.7995 \\
    \quad\(\hookrightarrow\) \(\Delta\)
    & & & \blue{-0.13}/\blue{-0.0004}
    & \blue{-0.22}/\red{+0.0006}
    & \red{+0.02}/\red{+0.0007}
    & \blue{-0.04}/\red{+0.0001}
    & \red{+0.04}/\red{+0.0002}
    & \red{+0.01}/\red{+0.0002}
    & \red{+0.01}/\red{+0.0004} \\

    \bottomrule
    \end{tabular}}
  }
\end{table*}
\begin{table*}[t]
  \centering
  \caption{Quantitative comparison (PSNR/SSIM) on benchmark datasets trained on DFLIP datasets~\cite{DIV2K, DF2KDataset, LSDIR, DiverSeg}. $^{\dagger}$ denotes the method is trained using $96\times 96$ patches. SST-L+ rows are shown in gray as reference results and are excluded from bolding. The best result among non-reference rows is bolded.}
  \label{tab:parameter_scale}

  {\renewcommand{\arraystretch}{1}
  \resizebox{\textwidth}{!}{%
    \begin{tabular}{@{} l c c *{7}{c} @{}}
    \toprule
    \multirow{2}{*}{Method} &
    \multirow{2}{*}{scale} &
    \multirow{2}{*}{\#params} &
    \multicolumn{1}{c}{Set5} &
    \multicolumn{1}{c}{Set14} &
    \multicolumn{1}{c}{BSD100} &
    \multicolumn{1}{c}{Urban100} &
    \multicolumn{1}{c}{Manga109} &
    \multicolumn{1}{c}{DIV2K-val} &
    \multicolumn{1}{c}{LSDIR-val} \\
    \cmidrule(lr){4-10}
    & & & PSNR/SSIM & PSNR/SSIM & PSNR/SSIM & PSNR/SSIM & PSNR/SSIM & PSNR/SSIM & PSNR/SSIM \\
    \midrule

    MambaIRV2-L~\cite{MambaIRV2}   & \multirow{3}{*}{$\times 2$} & 34.2 & \textbf{38.77}/\textbf{0.9636} & 35.20/0.9287 & 32.74/\textbf{0.9065} & 35.20/0.9508 & 40.82/0.9820 & 37.16/0.9524 & 32.93/0.9321 \\
    \graycell{SST-L+}  &                             & \graycell{20.3} & \graycell{38.78/0.9637} & \graycell{35.28/0.9288} & \graycell{32.73/0.9064} & \graycell{35.63/0.9530} & \graycell{40.86/0.9821} & \graycell{37.32/0.9535} & \graycell{33.12/0.9338} \\
    SST-XL+                        &                             & 33.9 & \textbf{38.77}/\textbf{0.9636} & \textbf{35.40}/\textbf{0.9289} & \textbf{32.75}/0.9063 & \textbf{35.83}/\textbf{0.9540} & \textbf{41.01}/\textbf{0.9825} & \textbf{37.40}/\textbf{0.9539} & \textbf{33.24}/\textbf{0.9347} \\
    \midrule

    MambaIRV2-L~\cite{MambaIRV2}   & \multirow{3}{*}{$\times 3$} & 34.2 & 35.23/0.9341 & 31.56/\textbf{0.8586} & 29.66/0.8195 & 31.13/0.9000 & 35.98/0.9574 & 33.43/0.9020 & 29.23/0.8586 \\
    \graycell{SST-L+}  &                             & \graycell{20.7} & \graycell{35.33/0.9345} & \graycell{31.54/0.8576} & \graycell{29.64/0.8193} & \graycell{31.53/0.9046} & \graycell{36.12/0.9580} & \graycell{33.55/0.9037} & \graycell{29.33/0.8609} \\
    SST-XL+                        &                             & 34.3 & \textbf{35.39}/\textbf{0.9350} & \textbf{31.62}/0.8584 & \textbf{29.69}/\textbf{0.8204} & \textbf{31.82}/\textbf{0.9077} & \textbf{36.30}/\textbf{0.9589} & \textbf{33.64}/\textbf{0.9047} & \textbf{29.46}/\textbf{0.8631} \\
    \midrule

    MambaIRV2-L~\cite{MambaIRV2}   & \multirow{3}{*}{$\times 4$} & 34.2 & 33.20/0.9073 & 29.57/0.8018 & 28.11/\textbf{0.7556} & 28.75/0.8526 & 32.99/0.9334 & 31.40/0.8576 & 27.29/0.7967 \\
    \graycell{SST-L+}  &                             & \graycell{21.3} & \graycell{33.10/0.9067} & \graycell{29.58/0.8016} & \graycell{28.06/0.7541} & \graycell{29.06/0.8583} & \graycell{33.18/0.9345} & \graycell{31.50/0.8594} & \graycell{27.37/0.7991} \\
    SST-XL+                        &                             & 34.9 & \textbf{33.24}/\textbf{0.9078} & \textbf{29.66}/\textbf{0.8025} & \textbf{28.12}/0.7555 & \textbf{29.38}/\textbf{0.8636} & \textbf{33.40}/\textbf{0.9360} & \textbf{31.59}/\textbf{0.8608} & \textbf{27.50}/\textbf{0.8024} \\
    \bottomrule
    \end{tabular}}
  }
  \vspace{-0.15cm}
\end{table*}
\begin{table}[t]
\centering
\caption{
    Training and inference costs of attention variants on SST$+$.
    Statistics are measured as detailed in Table~\textcolor{red}{1}.
    $^\S$ indicates that additional zero padding was used because FlashAttention4~\cite{FlashAttn4} currently supports backward computation only for channel dimensions that are multiples of 32. 
}
\label{tab:AblationAttention_cost_only}
\begin{tabular}{ccccc}
\toprule
\multirow{2}{*}{\textbf{Attention}} &
\multicolumn{2}{c}{\textbf{Training}} &
\multicolumn{2}{c}{\textbf{Inference}} \\
\cmidrule(lr){2-3}\cmidrule(lr){4-5}
& sec/step & memory & latency & memory \\
\midrule
FlashAttention2~\cite{FlashAttn2} & 0.556 & 47.1GB & 503.7ms & 2825MB \\
FlashAttention3~\cite{FlashAttn3} & 0.479 & 47.1GB & 455.7ms & 2825MB \\
FlashAttention4~\cite{FlashAttn4} & 0.488$^\S$ & 48.1GB$^\S$ & 452.3ms & 2819MB \\
\bottomrule
\end{tabular}
\end{table}

\section{Licenses for the used assets}
Our experiments use only existing public research assets, and we do not introduce or redistribute a new dataset. For training, we use DIV2K~\cite{DIV2K} and Flickr2K~\cite{DF2KDataset} as DF2K, and additionally use LSDIR~\cite{LSDIR} and DiverSeg-IP~\cite{DiverSeg} as part of the DFLIP training setting. DIV2K is made available for academic research purposes only, and the copyright of each image remains with its original owner. LSDIR is also made available for academic research purposes only, with images collected from the Internet and copyright retained by the original owners. DiverSeg-IP is provided for non-commercial research and educational purposes; the code in the DiverSeg repository is released under the MIT License, while the data follow the terms of their original sources. Flickr2K was collected by the EDSR/NTIRE2017 authors using the Flickr API; as we could not identify a separate explicit license for the Flickr2K image files, we use them only in the standard non-commercial academic research setting, citing the original source.

For evaluation, we use the standard SR benchmark datasets Set5~\cite{Set5}, Set14~\cite{Set14}, BSD100~\cite{BSD100}, Urban100~\cite{Urban100}, Manga109~\cite{Manga109}, DIV2K-val, and LSDIR-val. Set5 and Set14 are used for academic evaluation only. BSD100 is used under non-commercial research and educational terms. Urban100 consists of Flickr images and is distributed in commonly used dataset cards under CC BY 4.0; we use it only for benchmark evaluation and cite the original Urban100 paper. Manga109 is permitted solely for academic purposes, forbids redistribution to third parties, and requires attribution to the original manga authors when dataset images are reproduced in papers or videos. We use all datasets only for research training or evaluation, cite the corresponding dataset papers, and do not redistribute any dataset images.

For software, our implementation is based on PyTorch~\cite{PyTorch} and BasicSR~\cite{BasicSR}, and uses FlashAttention kernels~\cite{FlashAttn, FlashAttn2, FlashAttn3, FlashAttn4} for hardware-efficient attention computation. PyTorch is distributed under a BSD-style license, FlashAttention is released under the BSD 3-Clause License, and BasicSR is released under the Apache License 2.0. We preserve the corresponding copyright and license notices when using or releasing code derived from these libraries. FlexAttention~\cite{FlexAttn}, used only for comparison or ablation where applicable, is part of PyTorch and follows the PyTorch license.

\section{Boundary-Crop Sensitivity Analysis}\label{sec:boundary_analysis}
Although our SST show lower PSNR/SSIM on Set5, Set14, and BSD100 under the standard evaluation protocol, this trend can be partially attributed to an interaction between the boundary-cropping procedure and the spatial characteristics of our SST, rather than solely to reconstruction quality.
Specifically, standard SR evaluation excludes boundary pixels according to the upscaling factor before computing PSNR/SSIM~\cite{VDSR}. 
While this practice is useful for reducing ambiguity caused by degradation misalignment near image boundaries, it also removes regions where the effect of our design choices can be more visible.

As shown in Table~\ref{tab:crop_compare}, disabling the boundary crop changes the relative behavior of the evaluated methods.
For example, SST-L$+$ exhibits a smaller performance decrease than prior methods on Set5 and Set14, and shows improved performance on BSD100.
This result indicates that the standard crop can underemphasize certain properties of our SST, particularly its behavior near image boundaries, rather than suggesting that boundary-free evaluation is necessarily more favorable
We attribute this tendency to two aspects of the SST architecture.
First, for small LR inputs, especially at $\times$4, the large-window attention in our SST can cover most of the LR feature map, resulting in an almost-global receptive field.
Compared with local receptive-field-based methods, this property may encourage more spatially consistent restoration over the full image, treating boundaries and centers almost identically.
However, when boundary regions are excluded from evaluation, this potential advantage is less directly reflected in the reported PSNR/SSIM.
This observation suggests a trade-off: the model may not always maximize center-region pixel-wise PSNR under cropped evaluation, while still preserving competitive full-image reconstruction behavior and boundary consistency.
Second, the nearest-neighbor image skip connection provides a stable low-frequency reference near image boundaries, which may help the network predict residual details without relying excessively on extrapolated context.

\section{Extended Experiments on Parameter Scaling}
As demonstrated in several previous works, ViTs exhibit strong scalability, achieving consistent performance gains as the number of parameters increases~\cite{22BViT, DINOV3}.
To investigate whether the proposed SST also benefits from parameter scaling, we train a parameter-enlarged variant, denoted as SST-XL$+$, whose parameter size is comparable to that of MambaIRV2-L~\cite{MambaIRV2}, and evaluate its performance.
As shown in Table~\ref{tab:parameter_scale}, SST-XL$+$ achieves a remarkable improvement over MambaIRV2-L, outperforming MambaIRV2-L by 0.69\,dB in PSNR on Urban100$\times$3.
Furthermore, SST-XL$+$ improves upon SST-L$+$ across most evaluation metrics, demonstrating that the proposed SST remains effective under parameter scaling.

\begin{table}[t]
\centering
\caption{
    Latency and memory usage across GPUs. 
    We use FlashAttention3 for an H200 GPU and FlashAttention2 for RTX4090 and A6000 GPUs.
    We report the measured median speed when restoring a 1280$\times$720 image 10 times in FP32.
    Memory usage is measured using PyTorch's \texttt{max\_memory\_allocated} function.
}
\label{tab:gpu_latency_memory}
\begin{tabular}{lccccccc}
\toprule
\multirow{2}{*}{\textbf{Methods}} &
\multirow{2}{*}{\textbf{Scale}} &
\multicolumn{2}{c}{\textbf{RTX4090}} &
\multicolumn{2}{c}{\textbf{A6000}} &
\multicolumn{2}{c}{\textbf{H200}} \\
\cmidrule(lr){3-4}\cmidrule(lr){5-6}\cmidrule(lr){7-8}
& & latency & memory & latency & memory & latency & memory \\
\midrule
HAT & \multirow{9}{*}{$\times2$} & 1722.8ms & 9047MB & 3068.5ms & 9047MB & 709.7ms & 9070MB \\
ATD &  & 2665.8ms & 6216MB & 5064.8ms & 6216MB & 1266.8ms & 6239MB \\
PFT &  & OOM & OOM & 12990.1ms & 28847MB & 1920.8ms & 28869MB \\
MambaIR &  & 1570.8ms & 6883MB & 3349.4ms & 7680MB & 1043.8ms & 7704MB \\
MambaIRV2 &  & 3373.6ms & 5749MB & 9814.8ms & 5792MB & 2308.8ms & 5816MB \\
\cmidrule(lr){1-1}\cmidrule(lr){3-8}
SST & & 956.7ms & 2675MB & 1996.8ms & 2675MB & 428.9ms & 2675MB \\
SST$+$ & & 1046.0ms & 2825MB & 2125.3ms & 2825MB & 455.8ms & 2825MB \\
SST-L &  & 1439.2ms & 2836MB & 3051.8ms & 2824MB & 608.4ms & 2824MB \\
SST-L$+$ & & 1552.7ms & 2990MB & 3212.9ms & 2990MB & 644.9ms & 2990MB \\
\midrule
HAT & \multirow{9}{*}{$\times3$} & 799.8ms & 4040MB & 1377.0ms & 4040MB & 317.8ms & 4063MB \\
ATD &  & 1210.3ms & 2795MB & 2220.6ms & 2795MB & 550.0ms & 2819MB \\
PFT &  & 2088.5ms & 13525MB & 4646.3ms & 13525MB & 892.7ms & 13549MB \\
MambaIR &  & 668.4ms & 3125MB & 1358.8ms & 3458MB & 465.1ms & 3481MB \\
MambaIRV2 &  & 1434.5ms & 2589MB & 4161.6ms & 2632MB & 1000.3ms & 2655MB \\
\cmidrule(lr){1-1}\cmidrule(lr){3-8}
SST &  & 401.1ms & 1296MB & 739.2ms & 1296MB & 194.2ms & 1296MB \\
SST$+$ &  & 446.8ms & 1527MB & 799.9ms & 1527MB & 207.1ms & 1527MB \\
SST-L &  & 609.1ms & 1404MB & 1112.5ms & 1391MB & 276.3ms & 1391MB \\
SST-L$+$ & & 667.0ms & 1639MB & 1191.6ms & 1639MB & 293.4ms & 1639MB \\
\midrule
HAT & \multirow{9}{*}{$\times4$} & 510.8ms & 2428MB & 821.3ms & 2428MB & 195.8ms & 2452MB \\
ATD &  & 756.1ms & 1692MB & 1248.9ms & 1692MB & 339.0ms & 1716MB \\
PFT &  & 1237.4ms & 7302MB & 3168.1ms & 7302MB & 493.3ms & 7326MB \\
MambaIR &  & 356.5ms & 1818MB & 733.9ms & 1990MB & 269.7ms & 2014MB \\
MambaIRV2 &  & 848.9ms & 1574MB & 2188.1ms & 1616MB & 604.5ms & 1640MB \\
\cmidrule(lr){1-1}\cmidrule(lr){3-8}
SST &  & 199.4ms & 764MB & 401.9ms & 764MB & 111.5ms & 764MB \\
SST$+$ &  & 228.9ms & 914MB & 420.6ms & 914MB & 120.8ms & 914MB \\
SST-L &  & 308.6ms & 907MB & 556.3ms & 919MB & 158.3ms & 941MB \\
SST-L$+$ &  & 346.0ms & 1003MB & 613.1ms & 1003MB & 170.4ms & 1015MB \\
\midrule
\bottomrule
\end{tabular}
\end{table}

\section{Comparisons of Training and Inference Costs Across FlashAttention Implementations}
In this section, we compare the training and inference costs across different FlashAttention implementations, including FlashAttention2~\cite{FlashAttn2}, FlashAttention3~\cite{FlashAttn3}, and FlashAttention4~\cite{FlashAttn4}.
Since FlashAttention4 is still in beta and is primarily optimized to address the architectural characteristics and performance bottlenecks of NVIDIA Blackwell GPUs~(e.g, B200), we mainly use FlashAttention3 in our experiments.
As shown in Table~\ref{tab:AblationAttention_cost_only}, the inference cost consistently decreases as the FlashAttention implementation advances across versions.
These results suggest that, as FlashAttention kernels and GPU architectures continue to evolve, our method can further benefit from such system-level improvements with only minimal modifications to the proposed method itself.

\begin{table}[t]
\caption{
    Comparisons of lightweight SR methods trained on the DIV2K dataset.
    We reuse the statistics reported in prior work~\cite{ESC} and additionally include results we measured, following their procedure.
    The best result is bolded.
}\label{tab:light_div2k}
\resizebox{\textwidth}{!}{%
\begin{tabular}{@{}l|c|cccc|ccccc@{}}
\toprule
\multirow{2}{*}{Method} & \multirow{2}{*}{Scale} & \multirow{2}{*}{\begin{tabular}[c]{@{}c@{}}Latency\\ (ms)\end{tabular}} & \multirow{2}{*}{\begin{tabular}[c]{@{}c@{}}Mem\\ (MB)\end{tabular}} & \multirow{2}{*}{\begin{tabular}[c]{@{}c@{}}\#FLOPs\\ (G)\end{tabular}} & \multirow{2}{*}{\begin{tabular}[c]{@{}c@{}}\#params\\ (K)\end{tabular}} & \multicolumn{5}{c}{PSNR / SSIM} \\
 &  &  &  &  &  & Set5 & Set14 & B100 & Urban100 & Manga109 \\ \midrule
SwinIR-lt~\cite{SwinIR}& \multirow{14}{*}{$\times2$} & 1409.8 & 1287 & 244.2 & 910 & 38.14/0.9611 & 33.86/0.9206 & 32.31/0.9012 & 32.76/0.9340 & 39.12/0.9783 \\
ELAN-lt~\cite{ELAN} &  & 94.5 & 887 & 203.1 & 621 & 38.17/0.9611 & 33.94/0.9207 & 32.30/0.9012 & 32.76/0.9340 & 39.11/0.9782 \\
OmniSR~\cite{OmniSR} &  & 120.3 & 1031 & 194.5 & 772 & 38.22/0.9613 & 33.98/0.9210 & 32.36/0.9020 & 33.05/0.9363 & 39.28/0.9784 \\
SRFormer-lt~\cite{SRFormer} &  & 1456.3 & 1184 & 236.3 & 853 & 38.23/0.9613 & 33.94/0.9209 & 32.36/0.9019 & 32.91/0.9353 & 39.28/0.9785 \\
ATD-lt~\cite{ATD} &  & 733.5 & 2839 & 380.0 & 753 & 38.29/0.9616 & 34.10/0.9217 & 32.39/0.9023 & 33.27/0.9375 & 39.52/0.9789 \\
HiT-SRF~\cite{HiTSR} &  & 268.1 & 1804 & 226.5 & 847 & 38.26/0.9615 & 34.01/0.9214 & 32.37/0.9023 & 33.13/0.9372 & 39.47/0.9787 \\
ASID-D8~\cite{ASID} &  & 131.2 & 999 & 190.5 & 732 & 38.32/0.9618 & 34.24/0.9232 & 32.40/0.9028 & 33.35/0.9387 & -~/~- \\
MambaIR-lt~\cite{MambaIR} &  & 277.1 & 1695 & 334.2 & 905 & 38.13/0.9610 & 33.95/0.9208 & 32.31/0.9013 & 32.85/0.9349 & 39.20/0.9782 \\
MambaIRV2-lt~\cite{MambaIRV2} &  & 580.4 & 2824 & 286.3 & 774 & 38.26/0.9615 & 34.09/0.9221 & 32.36/0.9019 & 33.26/0.9378 & 39.35/0.9785 \\
RDN~\cite{RDN} &  & 279.3 & 2058 & 5096.2 & 22123 & 38.24/0.9614 & 34.01/0.9212 & 32.34/0.9017 & 32.89/0.9353 & 39.18/0.9780 \\
RCAN~\cite{RCAN} &  & 299.3 & 626 & 3529.7 & 15445 & 38.27/0.9614 & 34.12/0.9216 & 32.41/0.9027 & 33.34/0.9384 & 39.44/0.9786 \\
ESC~\cite{ESC} &  & 120.9 & 831 & 592.0 & 947 & \textbf{38.35}/0.9619 & 34.11/0.9223 & 32.41/0.9027 & 33.46/0.9395 & 39.54/0.9790 \\
UCAN~\cite{UCAN} &  & 345.3 & 6844 & 146.3 & 689 & 38.34/0.9618 & \textbf{34.27}/\textbf{0.9242} & 32.39/0.9025 & 33.22/0.9379 & 39.54/0.9790 \\
\textbf{SST-lt~(Ours)} &  & 191.9 & 755 & 2057.6 & 893 & \textbf{38.35}/\textbf{0.9620} & 34.14/0.9227 & 32.41/0.9028 & 33.54/0.9399 & 39.48/0.9785 \\
\textbf{SST-lt$+$~(Ours)} &  & 249.6 & 804 & 4903.3 & 893 & \textbf{38.35}/0.9619 & 34.22/0.9232 & \textbf{32.44}/\textbf{0.9031} & \textbf{33.79}/\textbf{0.9417} & \textbf{39.59}/\textbf{0.9788} \\\midrule
SwinIR-lt~\cite{SwinIR}& \multirow{14}{*}{$\times3$} & 331.7 & 596 & 110.8 & 918 & 34.62/0.9289 & 30.54/0.8463 & 29.20/0.8082 & 28.66/0.8624 & 33.98/0.9478 \\
ELAN-lt~\cite{ELAN} &  & 32.5 & 399 & 90.1 & 629 & 34.61/0.9288 & 30.55/0.8463 & 29.21/0.8081 & 28.69/0.8624 & 34.00/0.9478 \\
OmniSR~\cite{OmniSR} &  & 41.2 & 476 & 88.4 & 780 & 34.70/0.9294 & 30.57/0.8469 & 29.28/0.8094 & 28.84/0.8656 & 34.22/0.9487 \\
SRFormer-lt~\cite{SRFormer} &  & 530.5 & 537 & 105.4 & 861 & 34.67/0.9296 & 30.57/0.8469 & 29.26/0.8099 & 28.81/0.8655 & 34.19/0.9489 \\
ATD-lt~\cite{ATD} &  & 274.4 & 1258 & 168.0 & 760 & 34.74/0.9300 & 30.68/0.8485 & 29.32/0.8109 & 29.17/0.8709 & 34.60/0.9506 \\
HiT-SRF~\cite{HiTSR} &  & 124.9 & 1464 & 101.6 & 855 & 34.75/0.9300 & 30.61/0.8475 & 29.29/0.8106 & 28.99/0.8687 & 34.53/0.9502 \\
ASID-D8~\cite{ASID} &  & 61.9 & 460 & 86.4 & 739 & \textbf{34.84}/0.9307 & 30.66/0.8491 & 29.32/0.8119 & 29.08/0.8706 & -~/~- \\
MambaIR-lt~\cite{MambaIR} &  & 109.3 & 760 & 148.5 & 913 & 34.63/0.9288 & 30.54/0.8459 & 29.23/0.8084 & 28.70/0.8631 & 34.12/0.9479 \\
MambaIRV2-lt~\cite{MambaIRV2} &  & 259.0 & 1250 & 126.7 & 781 & 34.71/0.9298 & 30.68/0.8483 & 29.26/0.8098 & 29.01/0.8689 & 34.41/0.9497 \\
RDN~\cite{RDN} &  & 146.5 & 985 & 2281.2 & 22308 & 34.71/0.9296 & 30.57/0.8468 & 29.26/0.8093 & 28.80/0.8653  & 34.13/0.9484 \\
RCAN~\cite{RCAN} &  & 85.1 & 560 & 1586.1 & 15629 & 34.74/0.9299 & 30.65/0.8482 & 29.32/0.8111 & 29.09/0.8702 & 34.44/0.9499 \\
ESC~\cite{ESC} &  & 41.4 & 385 & 267.6 & 955 & \textbf{34.84}/\textbf{0.9308} & \textbf{30.74}/0.8493 & \textbf{29.34}/0.8118 & 29.28/0.8739 & 34.66/0.9512 \\
UCAN~\cite{UCAN} &  & 143.7 & 3167 & 64.6 & 696 & 34.83/\textbf{0.9308} & 30.72/0.8493 & 29.32/0.8121 & 29.15/0.8712 & 34.62/0.9508 \\
\textbf{SST-lt~(Ours)} &  & 69.2 & 359 & 954.1 & 900 & 34.79/0.9305 & 30.68/0.8489 & 29.31/0.8113 & 29.34/0.8748 & 34.55/0.9507 \\
\textbf{SST-lt$+$~(Ours)} &  & 98.1 & 432 & 2397.7 & 900 & 34.80/\textbf{0.9308} & 30.71/\textbf{0.8495} & \textbf{29.34}/\textbf{0.8122} & \textbf{29.50}/\textbf{0.8774} & \textbf{34.67}/\textbf{0.9514} \\ \midrule
SwinIR-lt~\cite{SwinIR}& \multirow{14}{*}{$\times4$} & 222.9 & 351 & 63.6 & 930 & 32.44/0.8976 & 28.77/0.7858 & 27.69/0.7406 & 26.47/0.7980 & 30.92/0.9151 \\
ELAN-lt~\cite{ELAN} &  & 18.0 & 241 & 54.1 & 640 & 32.43/0.8975 & 28.78/0.7858 & 27.69/0.7406 & 26.54/0.7982 & 30.92/0.9150 \\
OmniSR~\cite{OmniSR} &  & 22.5 & 273 & 50.9 & 792 & 32.49/0.8988 & 28.78/0.7859 & 27.71/0.7415 & 26.64/0.8018 & 31.02/0.9151 \\
SRFormer-lt~\cite{SRFormer} &  & 287.2 & 329 & 62.8 & 873 & 32.51/0.8988 & 28.82/0.7872 & 27.73/0.7422 & 26.67/0.8032 & 31.17/0.9165 \\
ATD-lt~\cite{ATD} &  & 189.7 & 753 & 100.1 & 769 & 32.63/0.8998 & 28.89/0.7886 & 27.79/0.7440 & 26.97/0.8107 & 31.48/0.9198 \\
HiT-SRF~\cite{HiTSR} &  & 82.1 & 1331 & 58.0 & 866 & 32.55/0.8999 & 28.87/0.7880 & 27.75/0.7432 & 26.80/0.8069 & 31.26/0.9171 \\
ASID-D8~\cite{ASID} &  & 61.8 & 265 & 49.6 & 748 & 32.57/0.8990 & 28.89/0.7898 & 27.78/0.7449 & 26.89/0.8096 & -~/~- \\
MambaIR-lt~\cite{MambaIR} &  & 55.8 & 438 & 84.6 & 924 & 32.42/0.8977 & 28.74/0.7847 & 27.68/0.7400 & 26.52/0.7983 & 30.94/0.9135\\
MambaIRV2-lt~\cite{MambaIRV2} &  & 153.4 & 748 & 75.6 & 790 & 32.51/0.8992 & 28.84/0.7878 & 27.75/0.7426 & 26.82/0.8079 & 31.24/0.9182 \\
RDN~\cite{RDN} &  & 66.0 & 791 & 1309.2 & 22271 & 32.47/0.8990 & 28.81/0.7871 & 27.72/0.7419 & 26.61/0.8028 & 31.00/0.9151 \\
RCAN~\cite{RDN} &  & 52.2 & 540 & 917.6 & 15592 & 32.63/0.9002 & 28.87/0.7889 & 27.77/0.7436 & 26.82/0.8087 & 31.22/0.9173 \\
ESC~\cite{ESC} &  & 21.9 & 215 & 149.2 & 968 & \textbf{32.68}/\textbf{0.9011} & 28.93/0.7902 & 27.80/0.7447 & 27.07/0.8144 & 31.54/0.9207 \\
UCAN~\cite{UCAN} &  & 81.8 & 1780 & 38.1 & 702 & 32.65/0.9010 & \textbf{28.95}/0.7899 & 27.79/0.7454 & 26.89/0.8097 & 31.50/0.9200 \\
\textbf{SST-lt~(Ours)} &  & 36.5 & 205 & 517.0 & 908 & 32.62/0.9007 & 28.93/0.7896 & 27.79/0.7446 & 27.12/0.8158 & 31.48/0.9200 \\
\textbf{SST-lt$+$~(Ours)} &  & 53.0 & 253 & 1329.7 & 908 & 32.65/0.9003 & 28.94/\textbf{0.7908} & \textbf{27.81}/\textbf{0.7454} & \textbf{27.23}/\textbf{0.8195} & \textbf{31.63}/\textbf{0.9219} \\ \bottomrule
\end{tabular}%
}
\end{table}

\begin{table}[t]
\caption{
    Comparisons of lightweight SR methods trained on the DFLIP dataset.
    We reuse the statistics reported in prior work~\cite{ESC} and additionally include our results, measured following their procedure.
    The best result is bolded.
}\label{tab:light_dflip}
\resizebox{\textwidth}{!}{%
\begin{tabular}{@{}l|c|cccc|ccccc@{}}
\toprule
\multirow{2}{*}{Method} & \multirow{2}{*}{Scale} & \multirow{2}{*}{\begin{tabular}[c]{@{}c@{}}Latency\\ (ms)\end{tabular}} & \multirow{2}{*}{\begin{tabular}[c]{@{}c@{}}Mem\\ (MB)\end{tabular}} & \multirow{2}{*}{\begin{tabular}[c]{@{}c@{}}\#FLOPs\\ (G)\end{tabular}} & \multirow{2}{*}{\begin{tabular}[c]{@{}c@{}}\#params\\ (K)\end{tabular}} & \multicolumn{5}{c}{PSNR / SSIM} \\
 &  &  &  &  &  & Set5 & Set14 & B100 & Urban100 & Manga109 \\ \midrule
SRFormer-lt~\cite{SRFormer} & \multirow{6}{*}{$\times2$} & 1838.1 & 1184 & 236.3 & 853 & 38.24/0.9615 & 34.13/0.9218 & 32.42/0.9026 & 33.37/0.9386 & 39.36/0.9787 \\
ATD-lt~\cite{ATD} &  & 733.5 & 2839 & 380.0 & 753 & 38.29/0.9616 & 34.30/0.9230 & 32.43/0.9027 & 33.62/0.9401 & 39.60/0.9791 \\
HiT-SRF~\cite{HiTSR} &  & 268.1 & 1804 & 226.5 & 847 & 38.31/0.9616 & 34.31/0.9230 & 32.45/0.9031 & 33.58/0.9404 & 39.69/0.9793 \\
ESC~\cite{ESC} &  & 120.9 & 831 & 592.0 & 947 & 38.34/0.9618 & 34.42/0.9235 & 32.50/0.9036 & 33.86/0.9424 & 39.73/0.9795 \\
\textbf{SST-lt~(Ours)} &  & 191.9 & 755 & 2057.6 & 893 & 38.39/\textbf{0.9622} & 34.50/0.9245 & 32.50/0.9038 & 34.03/0.9436 & 39.72/0.9795 \\
\textbf{SST-lt$+$~(Ours)} &  & 249.6 & 804 & 4903.3 & 893 & \textbf{38.42}/\textbf{0.9622} & \textbf{34.62}/\textbf{0.9250} & \textbf{32.53}/\textbf{0.9042} & \textbf{34.31}/\textbf{0.9452} & \textbf{39.90}/\textbf{0.9798} \\ \midrule
SRFormer-lt~\cite{SRFormer} & \multirow{4}{*}{$\times3$} & 668.3 & 537 & 105.4 & 861 & 34.67/0.9297 & 30.75/0.8484 & 29.30/0.8108 & 29.10/0.8701 & 34.26/0.9498 \\
ATD-lt~\cite{ATD} &  & 274.4 & 1258 & 168.0 & 760 & 34.71/0.9300 & 30.77/0.8493 & 29.33/0.8116 & 29.42/0.8743 & 34.61/0.9509 \\
HiT-SRF~\cite{HiTSR} &  & 124.9 & 1464 & 101.6 & 855 & 34.69/0.9298 & 30.81/0.8493 & 29.32/0.8115 & 29.28/0.8729 & 34.72/0.9511 \\
ESC~\cite{ESC} &  & 41.4 & 385 & 267.6 & 955 & 34.85/0.9312 & \textbf{30.97}/0.8511 & \textbf{29.41}/0.8135 & 29.70/0.8799 & 34.94/0.9525 \\
\textbf{SST-lt~(Ours)} &  & 69.2 & 359 & 954.1 & 900 & 34.87/0.9310 & 30.92/0.8508 & 29.38/0.8131 & 29.74/0.8807 & 34.79/0.9520 \\
\textbf{SST-lt$+$~(Ours)} &  & 98.1 & 432 & 2397.7 & 900 & \textbf{34.93}/\textbf{0.9316} & \textbf{30.97}/\textbf{0.8514} & \textbf{29.41}/\textbf{0.8140} & \textbf{29.95}/\textbf{0.8840} & \textbf{34.99}/\textbf{0.9528} \\ \midrule
SRFormer-lt~\cite{SRFormer} & \multirow{4}{*}{$\times4$} & 327.8 & 329 & 62.8 & 873 & 32.49/0.8993 & 28.89/0.7887 & 27.76/0.7429 & 26.90/0.8086 & 31.25/0.9189 \\
ATD-lt~\cite{ATD} &  & 189.7 & 753 & 100.1 & 769 & 32.52/0.8995 & 28.93/0.7896 & 27.79/0.7443 & 27.18/0.8150 & 31.47/0.9208 \\
HiT-SRF~\cite{HiTSR} &  & 82.1 & 1331 & 58.0 & 866 & 32.55/0.8997 & 28.96/0.7897 & 27.77/0.7443 & 27.07/0.8130 & 31.59/0.9208 \\
ESC~\cite{ESC} &  & 21.9 & 215 & 149.2 & 968 & \textbf{32.79}/\textbf{0.9025} & \textbf{29.06}/\textbf{0.7927} & 27.85/0.7466 & 27.45/0.8229 & 31.87/0.9239 \\
\textbf{SST-lt~(Ours)} &  & 36.5 & 205 & 517.0 & 908 & 32.70/0.9017 & 29.03/0.7918 & 27.83/0.7459 & 27.41/0.8225 & 31.64/0.9224 \\
\textbf{SST-lt$+$~(Ours) }&  & 53.0 & 253 & 1329.7 & 908 & 32.71/0.9019 & 29.04/0.7921 & \textbf{27.86}/\textbf{0.7472} & \textbf{27.58}/\textbf{0.8270} & \textbf{31.92}/\textbf{0.9241} \\ \bottomrule
\end{tabular}%
}
\end{table}

\begin{table}[t]
\caption{
    Comparisons of lightweight SR methods trained on the DFLIP dataset.
    We reuse the statistics reported in prior work~\cite{ESC} and additionally include our results, measured following their procedure.
    The best result is bolded.
}\label{tab:escrib_gpu}
\resizebox{\textwidth}{!}{%
\begin{tabular}{@{}l|c|cccc|ccccc@{}}
\toprule
\multirow{2}{*}{Method} & \multirow{2}{*}{Scale} & \multirow{2}{*}{\begin{tabular}[c]{@{}c@{}}Latency\\ (ms)\end{tabular}} & \multirow{2}{*}{\begin{tabular}[c]{@{}c@{}}Mem\\ (MB)\end{tabular}} & \multirow{2}{*}{\begin{tabular}[c]{@{}c@{}}\#FLOPs\\ (G)\end{tabular}} & \multirow{2}{*}{\begin{tabular}[c]{@{}c@{}}\#params\\ (K)\end{tabular}} & \multicolumn{5}{c}{PSNR / SSIM} \\
 &  &  &  &  &  & Set5 & Set14 & B100 & Urban100 & Manga109 \\ \midrule

ESC~\cite{ESC} & \multirow{2}{*}{$\times2$} & 120.9 & 831 & 592.0 & 947 & 38.34/0.9618 & 34.42/0.9235 & 32.50/0.9036 & 33.86/0.9424 & 39.73/0.9795 \\
ESC\textbf{+RIB} &  & 121.0 & 901 & 1825.8 & 1008 & \textbf{38.40}/\textbf{0.9621} & \textbf{34.45}/\textbf{0.9236} & \textbf{32.52}/\textbf{0.9040} & \textbf{34.09}/\textbf{0.9440} & \textbf{39.77}/\textbf{0.9796} \\ \midrule

ESC~\cite{ESC} & \multirow{2}{*}{$\times3$} & 41.4 & 385 & 267.6 & 955 & \textbf{34.85}/0.9312 & 30.97/0.8511 & 29.41/0.8135 & 29.70/0.8799 & 34.94/0.9525 \\
ESC\textbf{+RIB} &  & 42.6 & 421 & 842.9 & 1016 & 34.83/\textbf{0.9313} & \textbf{30.98}/\textbf{0.8515} & \textbf{29.42}/\textbf{0.8140} & \textbf{29.88}/\textbf{0.8824} & \textbf{34.99}/\textbf{0.9529} \\ \midrule

ESC~\cite{ESC} & \multirow{2}{*}{$\times4$} & 21.9 & 215 & 149.2 & 968 & 32.79/0.9025 & \textbf{29.06}/\textbf{0.7927} & 27.85/0.7466 & 27.45/0.8229 & 31.87/0.9239 \\
ESC\textbf{+RIB} &  & 23.3 & 237 & 457.8 & 1029 & \textbf{32.82}/\textbf{0.9027} & \textbf{29.06}/0.7926 & \textbf{27.87}/\textbf{0.7474} & \textbf{27.59}/\textbf{0.8258} & \textbf{31.93}/\textbf{0.9245} \\ \bottomrule

\end{tabular}%
}
\end{table}
\begin{table}[!t]
\centering
\caption{Latency comparison on MacBook M2 Air. Statistics are measures while processing a 128$\times$128 input.}
\label{tab:escrib_mobile}
\begin{tabular}{lccc}
\toprule
Methods & Windows & FLOPs (G) & Latency (ms) \\
\midrule
ESC     & $32 \times 32$ & 41.3 & 181.16 \\
ESC\textbf{+RIB} & $64 \times 64$ & 123.8 & 142.52 \\
\bottomrule
\end{tabular}
\end{table}

\section{Comparisons of Inference Costs Across GPUs}
We then evaluate the inference efficiency of our approach against representative SR baselines~\cite{HAT, ATD, PFT, MambaIR, MambaIRV2} by measuring inference latency and memory usage on a range of GPUs.
We note that comparisons with RPB-based Transformers prior to HAT~\cite{HAT} are less meaningful, as they incur repetitive overhead from window-mask construction and RPB table-index generation. 
Therefore, we mainly compare against RPB-based Transformers built on the HAT codebase.
As summarized in Table~\ref{tab:gpu_latency_memory}, our SST variants consistently achieve substantially lower memory consumption and faster inference latency than competing methods across all tested GPUs, highlighting the practical efficiency of our FlashAttention-based implementation.

\section{Results on Lightweight Super-Resolution Tasks}
Although our primary goal is to improve performance through scaling, the practical advantages of FlashAttention also make our approach highly effective in lightweight regimes.
As shown in Tables~\ref{tab:light_div2k} and~\ref{tab:light_dflip}, our method substantially improves both efficiency and reconstruction quality over the existing RPB-based Transformers.
Notably, by leveraging scaling, our model achieves an impressive PSNR of 34.31\,dB on Urban100$\times$2 even under a parameter budget of fewer than 1M parameters.
These results highlight the practicality of our approach and demonstrate its potential for a wide range of applications.

\section{Applying RIB to Hybrid CNN-Transformer Architecture}
Our SST-light variants demonstrate clear advantages over lightweight RPB-based Transformers, achieving lower latency and higher performance.
However, SST-light variants show somewhat comparable results to hybrid CNN-Transformer architectures~\cite{ESC}, which combine convolutions that provide remarkable efficiency and competitive performance in small model sizes.
Nevertheless, we suggest that the proposed RIB is not limited to pure Transformers and can be orthogonally applied to hybrid CNN-Transformer architectures as well.
Specifically, we replace the RPB and 32$\times$32 window attention with FlexAttention in ESC with RIB and 64$\times$64 window attention using FlashAttention, thereby strengthening long-range interactions.
As demonstrated in Table~\ref{tab:escrib_gpu}, ESC enhanced with RIB achieves a remarkable 0.23\,dB PSNR improvement on Urban100$\times$2 while adding only a marginal amount of latency and memory usage over the original ESC.
These results demonstrate the versatility of RIB, suggesting its promise for hybrid CNN-Transformer architectures.
That said, the use of RIB with a larger window size substantially increases FLOPs, which could be a concern in the lightweight regime, where models are often expected to run efficiently even on compute-bound edge devices.
To verify whether self-attention with RIB remains robust in such environments, we measure the latency on a MacBook M2 Air.
As demonstrated in Table~\ref{tab:escrib_mobile}, ESC\textbf{+RIB} shows even lower latency despite requiring approximately 3$\times$ more FLOPs, indicating the potential effectiveness of RIB even on compute-bound edge devices.

\begin{figure}[t]
  \centering
  \includegraphics[width=\textwidth]{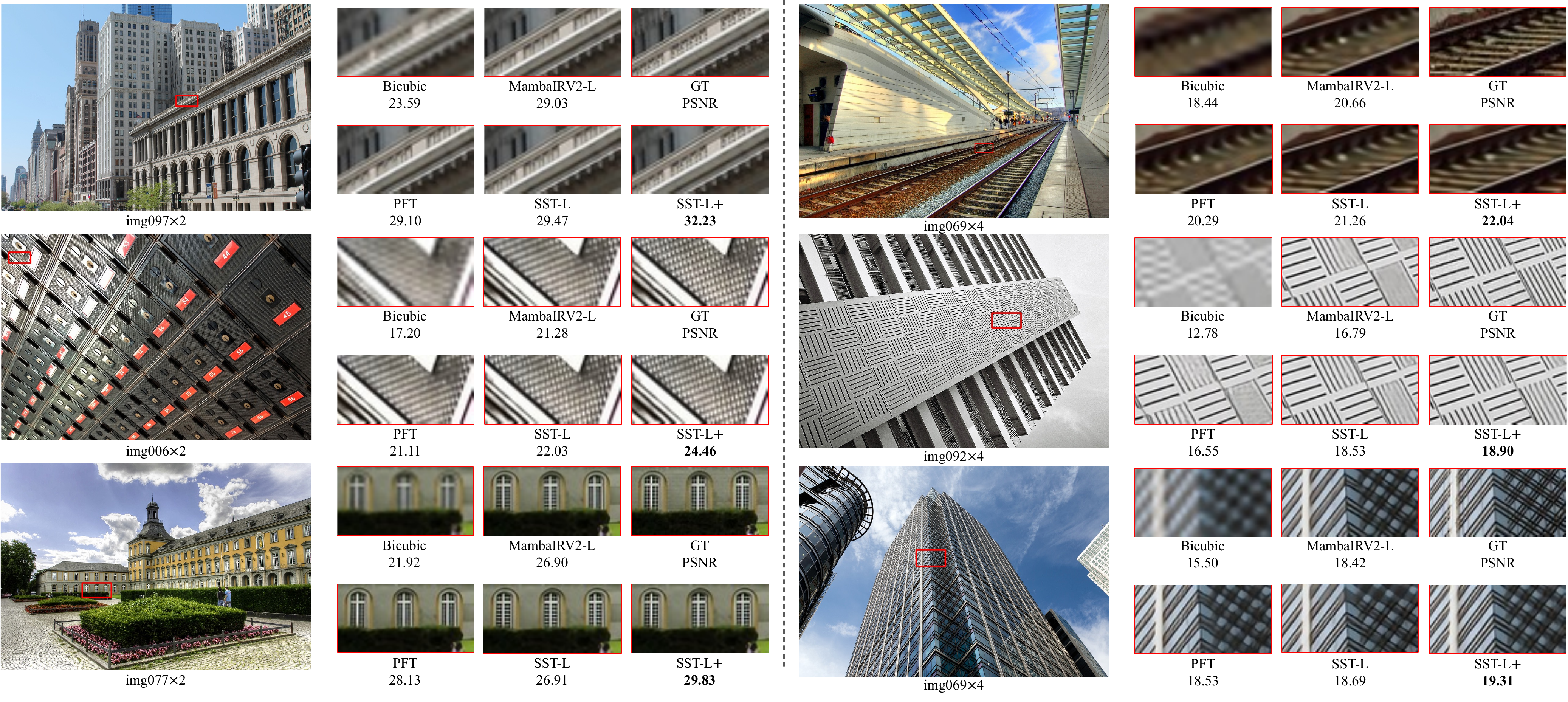}
  \caption{
    Visual comparison on an Urban100 dataset for scale $\times$2 and $\times$4.
  }
  \label{fig:Qualitative}
\end{figure}

\begin{figure}[t]
  \centering
  \includegraphics[width=\textwidth]{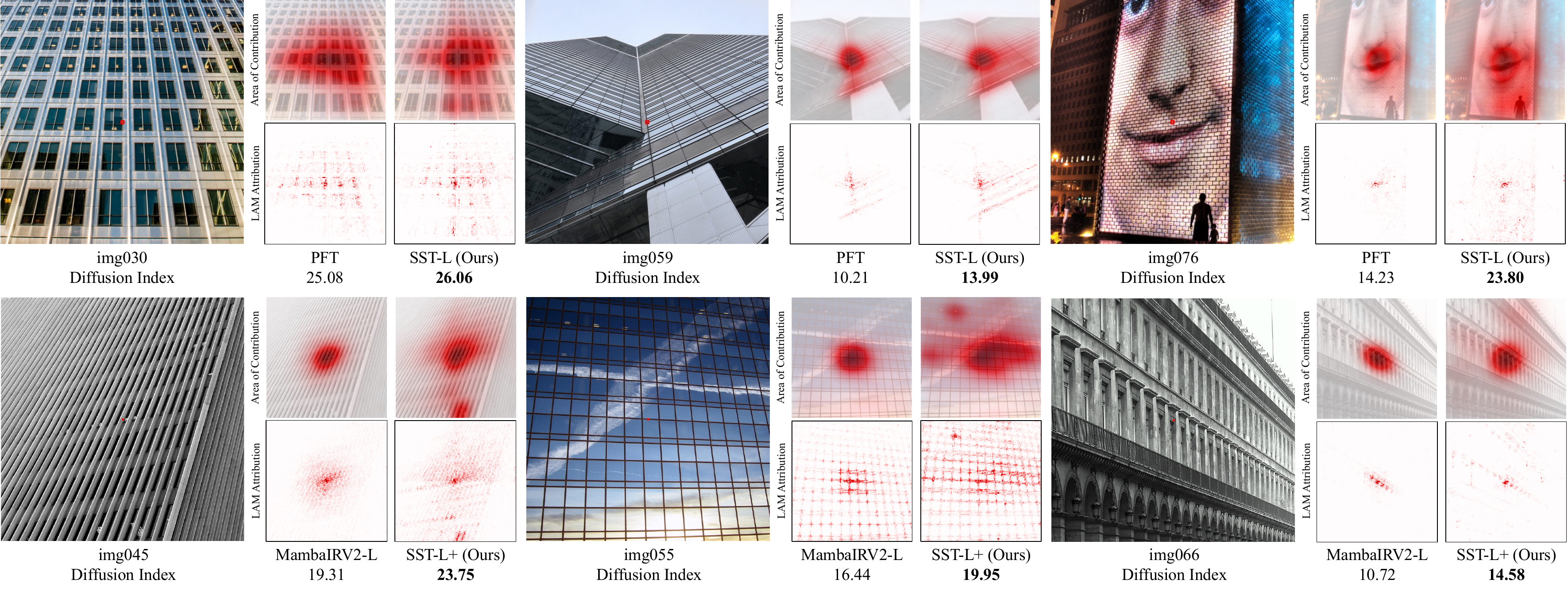}
  \caption{
    Local Attribution Map Analysis. Since padding and cropping for windowing inside networks hinder accurate comparison, we compare SST-L and SST-L$+$ separately, leveraging 512$\times$512 and 768$\times$768 images, respectively.  
  }
  \label{fig:LAM}
\end{figure}

\section{Visual Quality Comparisons and LAM Analyses}
In this section, we provide additional qualitative comparisons to complement the quantitative results in the main manuscript. As shown in Figure~\ref{fig:Qualitative}, our SST variants produce visually pleasing super-resolved images across challenging Urban100 examples. Compared with Bicubic, PFT, and MambaIRV2-L, our method more faithfully restores repetitive structures, sharp edges, and fine architectural details, while suppressing noticeable artifacts. These results demonstrate that the proposed scaling strategy not only improves PSNR but also leads to perceptually favorable reconstruction quality.

We further analyze the effective receptive behavior of different models using LAM analysis in Figure~\ref{fig:LAM}. The results show that our network consistently refers to a broader spatial region than the compared methods when reconstructing the target area. In particular, the larger attribution area and higher diffusion index indicate that our method can effectively exploit long-range contextual information. This confirms that the proposed large-window attention enabled by RIB allows the network to utilize the widest reference range among the compared models.

\section{Visual Comparisons on CLA and PWConv-only Gating}\label{sec:pca_cla}
\begin{figure}[t]
  \centering
  \includegraphics[width=\textwidth]{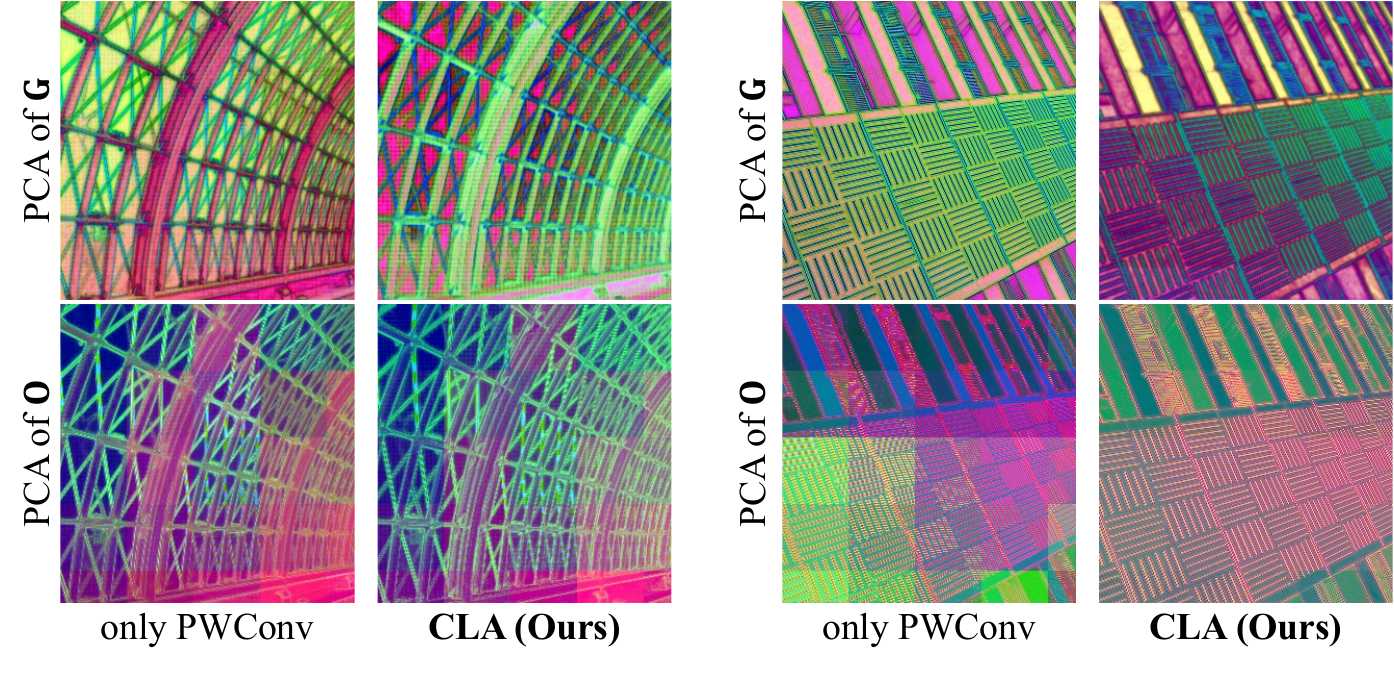}
  \caption{
    PCA visualization of the gating coefficient~($\mathbf{G}$) and self-attention output~($\mathbf{O}$) under PWConv and our proposed Convolutional Local Attention~(CLA).
    We visualize the first 64$\times$64 window self-attention within the last SST block.
    To visualize the extracted features, we apply Principal Component Analysis~(PCA) to reduce the channel dimension to three components and render them as RGB channels~\cite{SoMA}.
    Compared to PWConv-only variants, the self-attention layer with CLA is less affected by local textures and more robustly extracts structural features across the entire image.
  }
  \label{fig:CLACompare}
\end{figure}

In this section, we compare CLA with the PWConv-only gating module proposed in~\cite{GatedAttn} through qualitative visualizations.
As shown in Figure~\ref{fig:CLACompare}, when the gating module relies only on PWConv, the self-attention output~($\mathbf{O}$) is heavily influenced by local details, making it difficult to consistently capture repeated edge patterns. 
In contrast, when CLA is applied, the gating path~($\mathbf{G}$) takes responsibility for modeling local details, allowing $\mathbf{O}$ to focus more stably and robustly on repeated structures and edges.
These results suggest that, unlike conventional PWConv-only gating modules originally introduced for NLP tasks, CLA better reflects the characteristics required for the SR domain. 

\section{Limitations and Future Work}
Although RIB shows promising results for scaling SR Transformers, the scope of our current study is limited to SR. 
We believe SR is a particularly favorable setting for RIB because it involves a large number of dense pixel-level tokens, which carry relatively weak semantic abstraction, and because reconstruction can strongly benefit from matching repeated or near-identical local patterns such as textures, edges, and structures across distant regions. 
In this regime, preserving content similarity while injecting a decoupled geometric bias can provide a clearer advantage over RoPE, whose position-dependent rotation may suppress the similarity between repeated patterns at different spatial offsets.
However, this assumption does not necessarily hold for all high-resolution vision tasks. 
Tasks such as semantic segmentation, instance segmentation, object detection, and keypoint estimation often rely more on semantically rich tokens, object-level grouping, and category- or instance-level reasoning than on near-exact repeated-pattern matching. 
For such tasks, the benefit of RIB may be less direct, and a geometry-driven positional bias alone may not provide the same advantage observed in SR. 
Therefore, our results should be interpreted as evidence that RIB is effective for SR-like restoration settings, rather than as a general claim that RIB universally improves all high-resolution dense prediction tasks.
Nevertheless, we believe that extending RIB beyond classical SR remains an important direction for future work. 
In particular, investigating its applicability to other low-level vision and restoration tasks that share similar dense-token and repeated-pattern matching characteristics would be a valuable next step.

\end{document}